\documentclass[10pt,journal,compsoc]{IEEEtran}
\ifCLASSOPTIONcompsoc
  \usepackage[nocompress]{cite}
\else
  \usepackage{cite}
\fi

\usepackage{graphicx}

\usepackage{float}

\usepackage{tabularx}

	\usepackage[document]{ragged2e}   %% JUSTIFY TEXT --- ABSTRACT
	
\usepackage{makecell}

\usepackage{multicol}
 \usepackage{booktabs}
 \usepackage{multirow}

\usepackage{amssymb}% http://ctan.org/pkg/amssymb
\usepackage{pifont}% http://ctan.org/pkg/pifont

\usepackage[T2A,T1]{fontenc}
\usepackage[utf8]{inputenc}
\usepackage[russian,english]{babel}

\ifCLASSINFOpdf
\else
\fi
\begin{document}
\title{Data augmentation with automated machine learning: approaches and performance comparison with classical data augmentation methods }

\author{Alhassan~Mumuni$^1$$^*$~\IEEEmembership{}
		and~Fuseini~Mumuni$^2$~\IEEEmembership{}% <-this % stops a space
	\IEEEcompsocitemizethanks{\IEEEcompsocthanksitem $^1$Alhassan Mumuni:
		Department
		of Electrical and Electronics Engineering, Cape Coast Technical University, Cape Coast, Ghana.\protect\\
		$^*$Corresponding author, E-mail: alhassan.mumuni@cctu.edu.gh
		\IEEEcompsocthanksitem $^2$Fuseini Mumuni: University of Mines and Technology, UMaT, Tarkwa, Ghana. E-mail: fmumuni@umat.edu.gh}% <-this % stops an unwanted space
     }


\markboth{}%
{Shell \MakeLowercase{\textit{et al.}}: Bare Demo of IEEEtran.cls for Computer Society Journals}



\IEEEtitleabstractindextext{%
\begin{abstract}

\justify
		
Data augmentation is arguably the most important regularization technique commonly used to improve generalization performance of machine learning models. It primarily involves the application of appropriate data transformation operations to create new data samples with desired properties. Despite its effectiveness, the process is often challenging because of the time-consuming trial and error procedures for creating and testing different candidate augmentations and their hyperparameters manually. State-of-the-art approaches are increasingly relying on automated machine learning (AutoML) principles. This work presents a comprehensive survey of AutoML-based data augmentation techniques. We discuss various approaches for accomplishing data augmentation with AutoML, including data manipulation, data integration and data synthesis techniques. The focus of this work is on image data augmentation methods. Nonetheless, we cover other data modalities, especially in cases where the specific data augmentations techniques being discussed are more suitable for these other modalities. For instance, since automated data integration methods are more suitable for tabular data, we cover tabular data in the discussion of data integration methods. The work also presents extensive discussion of techniques for accomplishing each of the major subtasks of the image data augmentation process: search space design, hyperparameter optimization and model evaluation. Finally, we carried out an extensive comparison and analysis of the performance of automated data augmentation techniques and state-of-the-art methods based on classical augmentation approaches. The results show that AutoML methods for data augmentation currently outperform state-of-the-art techniques based on conventional approaches.

\end{abstract}

}

\maketitle

\IEEEdisplaynontitleabstractindextext

% \fi
\IEEEpeerreviewmaketitle

\justify

\section{Introduction}

\subsection{Background}

Practical implementations of machine learning systems require large data samples to produce satisfactory results. Since data is often not available in sufficient quantities, regularization techniques are critical for achieving good performance. These techniques commonly entail tweaking the machine learning model configuration or applying data augmentation—a range of methods for extending the available data by applying appropriate transformations. The basic idea is to modify training datasets by applying suitable transformations in ways that increase the quantity, representation quality and variability of the original data.

 The most commonly used data augmentation techniques include geometric transformations – particularly, rotation, flipping, shearing and scaling–and photometric transformations such as color jittering, solarizaion, brightness, contrast  adjustment, noise addition,  denoising, and color space conversion.  Data augmentation can also involve creating completely new data from scratch \cite{kollias2023abaw,tabak2023towards}.  This approach can be useful when the training data for the target application is inaccessible \cite{murtaza2023synthetic}. Methods for synthetic data generation include explicitly creating samples with desired data distribution using computer graphics tools (\cite{tabak2023towards}) or algorithmically generating artificial data with the aid of special deep learning models (e.g., with techniques such as differential neural rendering \cite{kwon2023renderable,zhang2023beyond}. Neural style transfer \cite{da2023generating}, generative modeling techniques such as variational autoencoders (VAEs) \cite{feng2023variational} and generative adversarial networks (GANs) \cite{wang2023applications} have also been extensively used to generate synthetic data for training deep learning models. Another way to augment training data is by integrating existing data from several sources (e.g., see \cite{suri2021ember,bai2021atj,li2021data}). This is a useful and a more natural way to augment training data since the approach can leverage the large quantities of data available in various forms on the internet and other sources.

\subsection{Limitations of classical data augmentation approaches}

While classical data augmentation methods can significantly improve the predictive performance of machine learning models, they are typically characterized by laborious manual work. The process of generating and finding the best augmentations for a particular dataset or task is a combinatorial problem, which requires infinitely large number of permutations of different settings to be tested for a suitable method to be found. However, for a given dataset, the number of unique augmentations that can be obtained with manual effort is generally limited. Moreover, it is known that different types of augmentations work well for different machine learning tasks, and determining a suitable augmentation methods for a particular task is a nontrivial problem.

Moreover, approaches that improve generalization on one dataset may fail to transfer to other datasets. For instance, Lopes et al. \cite{lopes2019improving} demonstrated that CutOut \cite{devries2017improved} improves performance on CIFAR-10 but not on ImageNet dataset. Also,  Raileanu et al. \cite{raileanu2021automatic} argue that classical  data augmentation approaches do not readily work well for reinforcement learning (RL) tasks. Generative modeling techniques such as VAEs and GANs have shown promise in generating synthetic data to alleviate data shortage problems but they also suffer from overfitting when trained on insufficient data. 

Approaches based on GANs are also not guaranteed to produce good results even in cases where sufficiently large and rich datasets are available \cite{ravuri2019seeing}. Moreover, since generative modeling-based approaches constrain the distribution of generated samples in some way (e.g., by requiring some form of similarity of the target domain and generated samples), many potentially useful augmentations that do not satisfy these constrains may be missed. AutoML approaches avoid this limitation by solving the problem in a similarity-independent manner.

The foregoing discussion shows that it is extremely challenging to achieve optimal augmentation results using traditional data augmentation methods. Even in situations where restrictions on the range of augmentations achievable do not exist, it is still difficult to obtain good results without excessive trial and error work. Recently, automated algorithmic solutions have been proposed as a means of simplifying the cumbersome data augmentation process. For example, transferable AutoMl  (Tr-AutoML) \cite{xue2019transferable} is specifically designed to allow learned features to be transferable to novel task domains. The approach combines meta-learning and architecture search for feature extraction from multiple but related datasets. Cubuk et al. \cite{cubuk2019autoaugment} show that AutoML-based augmentations are also transferable to new datasets. This could potentially provide a means for automatically learning useful augmentations that are independent of datasets for different machine learning tasks.

\subsection{Automated machine learning and data augmentation}

\begin{figure*}[!htb]
 	\vspace {-1mm}
 	\centering
 	\includegraphics[width=1.0 \linewidth]{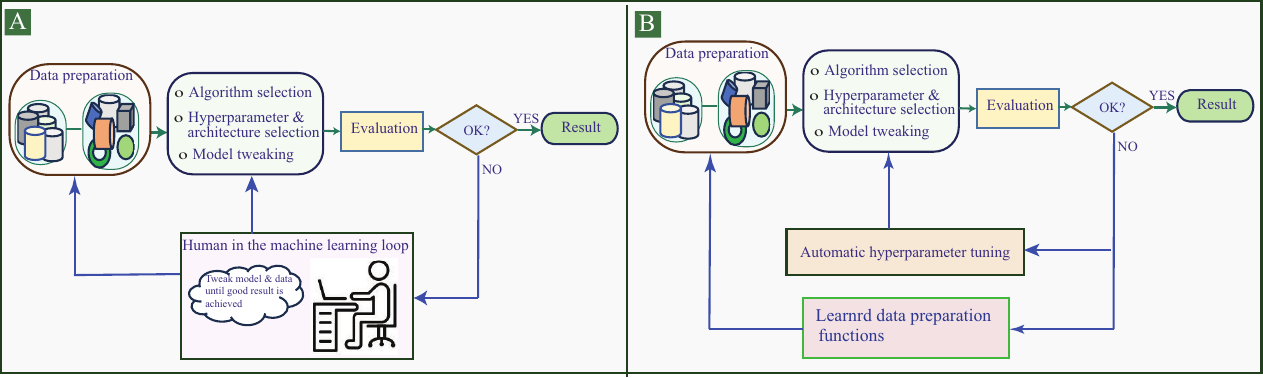} \vspace {-3mm}
 	\caption{Classical deep learning versus AutoML: In classical deep learning (A), all stages of the machine learning task—data preparation, hyperparameter selection and tuning, model selection and tweaking as well as the evaluation and validation of outcomes– are performed manually. In contrast, AutoML (B) incorporates an automatic tuning mechanism to learn the best parameters and hyperparameters for all these tasks. 
 	}\label{fig:1AutoML_vs_DL}
 \end{figure*}

Developing a machine learning model involves a series of tedious and repetitive tasks: data preparation, hyperparameter seclection, model selection, hyperparameter optimization, model tuining and evaluation of the resulting outcome. For a single machine learning tasks, all these steps are usually performed repeatedly until satisfactory results are achieved. This requires human experts to manually perform each of the subtasks (Figure \ref{fig:1AutoML_vs_DL} A). The task is inherently a combinatorial problem, as different types of hyperparameter settings would need to be configured and tested. Consequently, with the manual method it is often impractical to find an optimal solution due to the enormous variety of settings involved.

Automated machine learning (AutoML) \cite{kim2022survey} is an approach to automate all the processes of designing, training, deploying and monitoring machine learning solutions. AutoML frameworks can, for example, carry out data augmentation, perform additional processing and feature engineering, and construct the network structure of the machine learning model. The general idea of automated data augmentation is to create different types of basic transformations functions (e.g., rotations, flipping, color jittering, solarizaion, scaling, etc.) and then, using AutoML techniques, algorithmically apply various combinations of operations on the data and select the most effective set of data augmentation operations. Typically, black-box optimization techniques are used to find the best augmentation strategies. The search operation needs to find not only relevant transformations but also the optimal levels of transformations. For image augmentation these levels may be rotation angles, translation offsets and saturation values. Thus, automated data augmentation is mainly a task of combinatorial optimization of primitive data transformation operations. All these processes are carried out without the intervention of a human developer (see Figure \ref{fig:1AutoML_vs_DL} B). Given a task, dataset and performance objective, automated data augmentation produces optimal data augmentation policies that are applied on the dataset and trained in an end-to-end manner (i.e., from input data to final stage).
 
 %%%%%%%%%%%%%%%%% Figure 1 General AutoML vs DLS %%%%%%%%%%%%%%%%%%%% 

 %\usepackage{float}
 
 %%%%%%%%%%%%%%%%% End of Figure 1 %%%%%%%%%%%%%%%%%%%% 

\subsection{Motivation for this survey}

The concept of automated machine learning (AutoML) has received enormous attention in recent years. One of its most important applications is in data preparation, which includes data pre-processing and data augmentation. Despite there being many surveys on data augmentation methods \cite{shorten2019survey,khosla2020enhancing,mumuni2022data}, very few works have discussed automated data augmentation in sufficient detail. Among the many surveys on AutoML techniques, only a few  (e.g., \cite{waring2020automated,he2021automl,nagarajah2019review,tuggener2019automated,karmaker2021automl}) have discussed the application of AutoML to solve data augmentation problems. Unfortunately, the coverage of automated data augmentation in theses works is very limited in scope and depth. To the best of our knowledge, only Yang et al. \cite{yang2023survey} and Cheung \cite{cheung2023survey} have presented surveys that are specifically dedicated to automated data augmentation methods. However, these works do not cover many important aspects of AutoML approaches in data augmentation settings, including automated data integration and synthetic data generation methods. Moreover, approaches for the composition of augmentation functions as well as hyperparameter optimization strategies are not given sufficient attention. In addition, to the best of our knowledge, no work has conducted a detailed comparison of the predictive performance of AutoML-based data augmenation approaches and classical methods. This work has been motivated by the emerging importance of AutoML for data augmentation tasks and, as discussed above, the acute lack of coverage of many important issues in the literature.

\subsection{Main contributions}

The main contributions of this survey are that:

\begin{itemize}
	
	\item We discuss AutoML techniques for performing various data augmentation tasks, including data manipulation, data integration and data synthesis. The last two classes of tasks have not been covered in previous surveys.

	\item We present the main techniques for designing and composing transformation basic operations for automating data augmentation. We exhaustively describe the main characteristics, challenges  and workarounds of the various approaches.

	\item We extensively discuss a wide variety of search methods for finding optimal augmentation policies. The important concepts, properties as well as strengths and limitations of common black-box optimization methods are presented. In addition, we discuss alternative methods for obtaining effective augmentation policies without using these black-box optimization techniques.

	\item In addition, we present quantitative performance results of automated data augmentation approaches based on AutoML techniques. We additionally provide a comprehensive comparison of the predictive performance of automated data augmentation methods and classical approaches.
	
	 \item Finally, we discuss pertinent issues pertaining to the automation of data augmentation, and provide an overview of future research prospects.
	
\end{itemize}

%%%%%%%%%%%%%%%% FIGURE 2: Taxonomy   %%%%%%%%%%%%%%%%%%%%%%%%%%%%%%

%%%%%%%%%%%%%%%%% End of Figure 2 %%%%%%%%%%%%%%%%%%%%5

\subsection{Outline of survey}

The rest of the survey is organized as follows. Section 2 presents the basic concepts of AutoML and an overview of data augmentation in AutoML pipelines. We divide automated data augmentation tasks into three main subtasks: search space construction, optimization of augmentation policies and evaluation of learned strategies.  Section 3 presents a detailed discussion on three broad ways of achieving data augmentation, and discusses techniques for their realization in AutoML frameworks. Various techniques for construction of the search space are covered in Section 4. Here we discuss common methods for creating the set of data transformation operations or implicit parameters or neural models that can be used to augment data. The hyperparameter optimization subtask is discussed in Section 5. In Section 6, we present quantitative performance results that demonstrate the effectiveness of automated data augmentation techniques. We also compare the performance of these techniques with those of state-of-the-art classical data augmentation methods.  Section 7 discusses pertinent issues, unsolved problems and future research directions. Section 7 concludes the survey.

\section{Overview and general principle of AutoML-based data augmentation}

The task of AutoML-based data augmentation is to automatically generate the best augmentation for the given dataset and task. This is typically achieved by composing and applying a wide range of basic transformation operations on the given training data and then using various optimization techniques and heuristic search algorithms to find the most useful combination of augmentations.

%%%%%%%%%%%%%%%%% Figure2 General concept of automated data augmentation %%%%%%%%%%%%% 

%\usepackage{float}
\begin{figure*}[htb!]
	\vspace {-1mm}
	\centering
	\includegraphics[width=0.90 \linewidth]{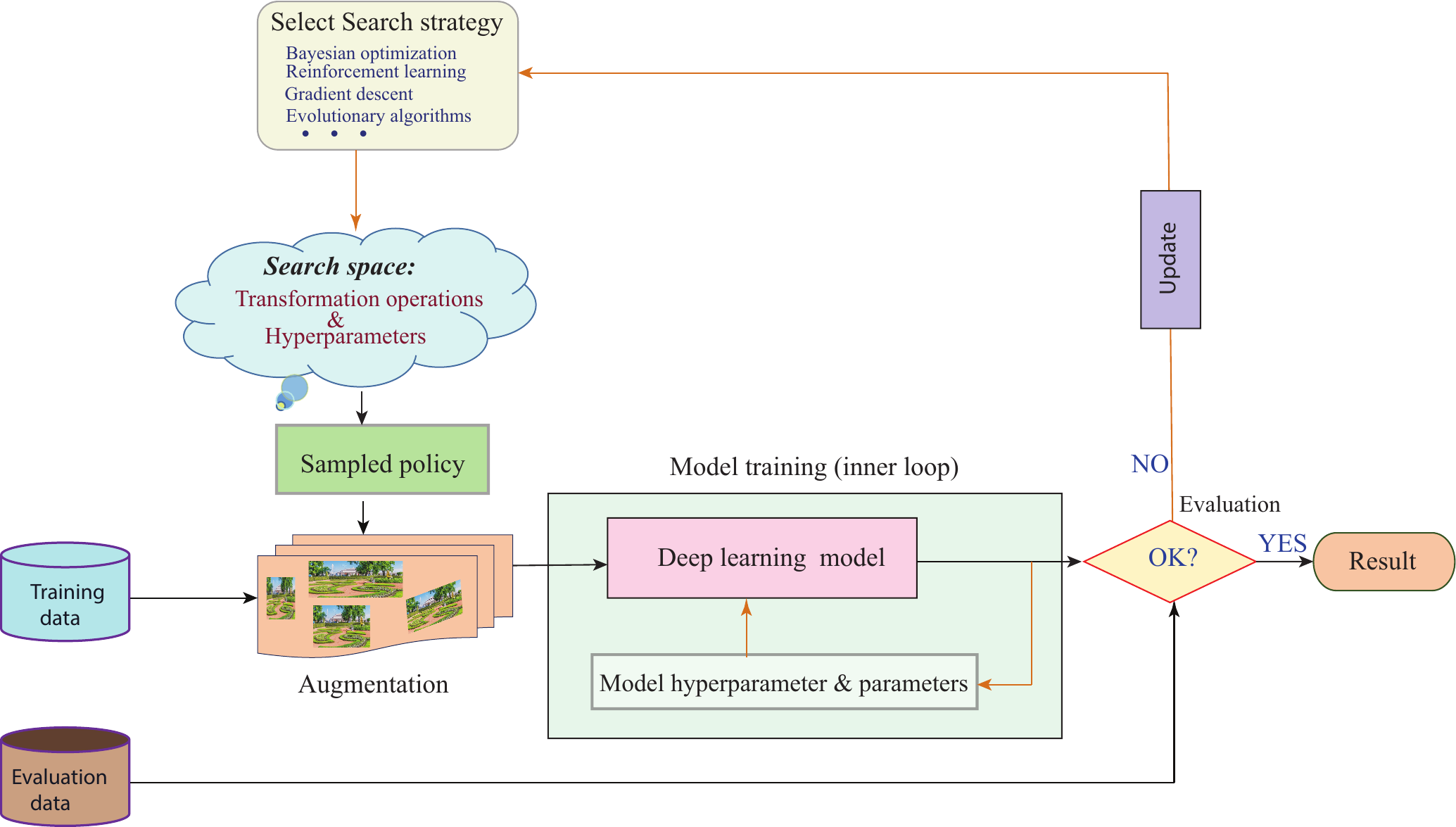} \vspace {-3mm}
	\caption{Bi-level optimization scheme and basic principle of operation of AutoML-based data augmentation methods. The general approach is to jointly optimize two machine learning loops – an outer loop involving augmentation hyperparameter search, and an inner loop that optimizes model parameters.
	}\label{fig:2BasicAtoML_DA}
\end{figure*}

%%%%%%%%%%%%%%%%% End of Figure 2 %%%%%%%%%%%%%%%%%%%% 

\subsection{General structure of AutoML-based data augmentation pipelines}
Automated data augmentation schemes typically utilize a bi-level optimization scheme ( see Figure \ref{fig:2BasicAtoML_DA}) in which an inner loop optimizes parameters of a deep neural network on training data while an outer loop optimizes the augmentation parameters based on the set of transformation operations and their associated constraints defined in the search space. For instance, the Automated augmentation scheme, proposed by Hataya et al. \cite{hataya2022meta}, simultaneously optimizes augmentation and image classification model parameters. With the proposed architecture, CNN parameters are optimized by minimizing the training loss while the tunable hyperparameters of the augmentation – transformation magnitude and the probability of application of the augmentation operations – are optimized by minimizing the validation loss.

\subsection{General procedure for AutoML-based data augmentation}
The first step in the AutoML-based data augmentation process is to define the relevant augmentations to incorporate. A good AutoML data augmentor should incorporate diverse operations that can generate a rich set of data to account for diverse real-world situations. The specification of the types and nature of augmentations typically involves defining different types of primitive data processing operations that transform data in desired ways (e.g., noise perturbation, rotation, scaling, blurring, contrast adjustment, etc.). In addition to the specific transformations, hyperparameters such as the degree or magnitudes of these transformations (e.g., the range of scale factors, rotation angels, etc.) are also specified. Another aspect of the augmentation problem is to specify how to combine these basic augmentations to compose more complex augmentations, known as augmentation policies. An augmentation policy is generally understood as a set of ordered transformation operations parameterized by selection probabilities and intensity values (see Figure \ref{fig:4AutoDA_Ops}). Optimization techniques and heuristic search algorithms are used to find the best augmentation policy. The best augmentation strategy is selected by evaluating the performance of candidate augmentations on the target or a proxy task according to a prior defined performance criterion. This task is generally formulated as a joint optimization problem, where performance on the end task is maximized by jointly optimizing model parameters, augmentation operations and their associated hyperparameters.  The most important factors in the choice of search strategy are the accuracy of the resulting policy, the search time and the cost in terms of computational resources.

Thus, the automated data augmentation tasks can generally be decomposed into three sub-problems:

\begin{enumerate}
	\item \textit{Search space composition}— This involves defining basic transformation operations and hyperparameters to be used for data augmentation. The types of augmentation operations depend on the data type and target task. For computer vision tasks, for example, these primitive augmentation operations are typically image transformation operations (e.g., affine and photometric transformations such as scaling, rotation, flipping,  shearing, color jittering, brightness adjustment and noise addition).   
	
 	\item \textit{Augmentation policy search}—This task, also known as augmentation policy optimization, is the task of formulating and applying search strategies to find the best combination of transformations and the associated hyperparameters in the search space that yield effective augmentations.
		
	\item \textit{Evaluation of performance of augmentation policies}—The final step in the automated data augmentation process is to evaluate the performance of all candidate policies and select the best. This is typically achieved by validating the performance of the resulting model on a proxy task or on the target task. It is common to solve tasks 2 and 3 as a single optimization problem.
	
\end{enumerate}

The different approaches for accomplishing each of these subtasks are thoroughly discussed in Sections 4 through 6. We present search space composition techniques-approaches for designing the set of data transformation operations or implicit parameters capable of learning transformations on data. Tunable hyperparameters that make it flexible to generate augmentations of varying magnitudes and with different attributes are also discussed. Heuristic search algorithms for finding the most effective transformations and their associated hyperparameters are covered in Section 5. Because in most implementations the evaluation strategies are realized as part of the optimization step, we only briefly cover this subtask (in Section 6).

\section{Approaches for data augmentation}

 \subsection{Data manipulation approaches}
\begin{figure*}[!htb]
 	\vspace {-2mm}
 	\centering
 	\includegraphics[width=0.85 \linewidth]{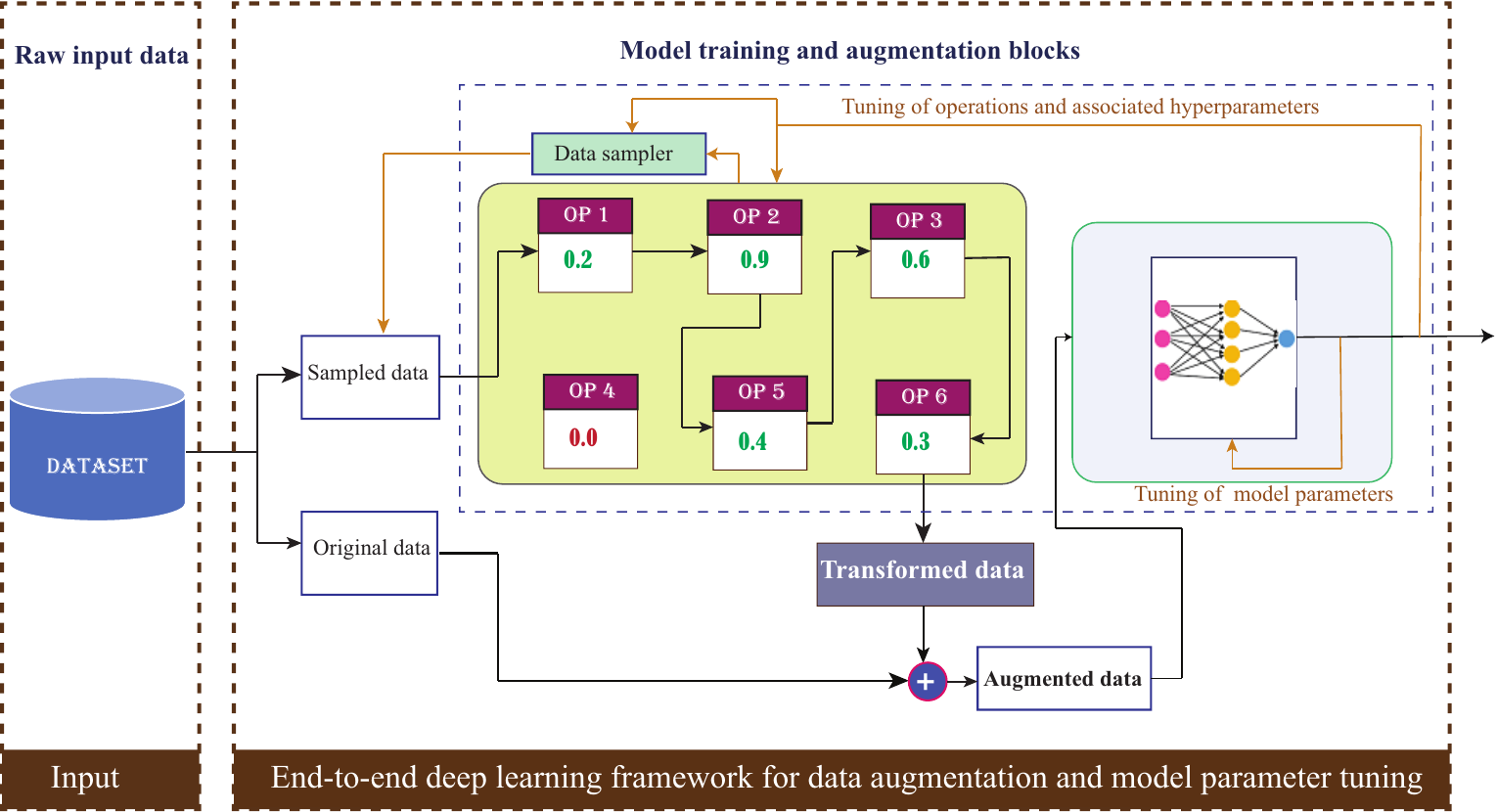} \vspace {-3mm}
 	\caption{General process of data manipulation-based augmentation using AutoML pipelines. With this approach, a subset of the input data is sampled by a learned sampler for transformation by different candidate augmentation policies. Different augmentation outcomes are produced by varying the ordering and application probabilities of the transformation functions OP1, OP2, .. OPn. 
 	}\label{fig:4AutoDA_Ops}
 \end{figure*}

 The easiest and most common way to extend training data is to manipulate existing data by applying appropriate transformations. With this method, typically, a large set of transformation operations, together with possible magnitudes and application probabilities are defined (see Figure \ref{fig:4AutoDA_Ops}). There are many ways to generate these basic transformation functions. They can be explicitly specified (e.g., \cite{cubuk2019autoaugment,hataya2020faster,li2020differentiable} or learned (e.g., \cite{mounsaveng2021learning,miaomulti}). In essence, this approach consists in designing a set of data transformation operations or implicit parameters or neural models capable of learning transformations on data. 
 
 \subsubsection{Image data manipulation}
 
 For automated image augmentation (e.g., \cite{cubukpractical,kashima2020joint,lin2021local}) the search space typically consists of a combination of spatial image transformation operations such as rotation, sheer, vertical and horizontal flipping, cropping and scaling, as well as photometric transformations like smoothing, solarization, sharpening, blurring brightness and contrast adjustment, noise addition, and color space conversion operations. Input data samples are augmented by different policies selected according to specific search criteria. In most cases, a policy in turn consists of sub-policies or proxy policies each of which applies a set of transformation operations on the training data in a sequential fashion.

 %%%%%%%%%%%%%%%% FIGURE 3: Search space condruction %%%%%%%%%%%%%%%%%%%%%%
 %\usepackage{float}
 %%%%%%%%%%%%%%%%% End of Figure 4 %%%%%%%%%%%%%%%%%%%%
 
 While this approach is widely successful, care must always be exercised in its application; the technique is aimed at changing the very nature of the training data. In particular, the performance of models trained on the resulting data is sensitive to not only the type of transformation operations, but also to the parameters of transformation functions. Consequently, incorrectly specified transformations or aggressive application of operations will often lead to unintended distortions that harm performance. This problem has been reported in several studies, including in the semial works AutoAugment \cite{cubuk2019autoaugment} and Population-Based Augmentation \cite{ho2019population}.  
 Because AutoML-based data augmentation methods based on data manipulation techniques have been extensively covered by other surveys (e.g., \cite{yang2023survey,cheung2023survey,xu2023comprehensive}), we rather focus on AutoML-based data augmentation strategies that have not received much attention. The most prominent of these approaches are dataset integration and data synthesis methods. These are covered in the next subsections.  
 
 \subsubsection{Text data manipulation}
 Manipulation-based text augmentation creates variations in the original text while still maintaining the meaning and grammatical structure. Approaches for manipulating textual data include rephrasing, randomly deleting or inserting words, replacing words by their corresponding synonyms, rearranging sentences in long texts, and introducing noise in the form of typographical errors. Natural language processing techniques \cite{niu2019automatically} are typically used to apply these manipulation methods in AutoML pipelines. Instead of explicitly performing transformations in the AutoML pipeline using basic NLP algorithms, approaches such as Text AutoAugment \cite{ren2021text} relies on learning text manipulation polices from input data. More recently, large language models have been seen as a vital tool for automated text augmentation \cite{dai2023auggpt,zhao2024improving}. Besides manipulating text, they can also perform a number of high-level functions such as model configuration or serving as an interface for interaction with the underlying AutoML framework (see \cite{tornede2023automl}).

 \subsubsection{Tabular data manipulation}
 Similar to image and text modalities, data manipulation operations can be applied on tables to directly transform existing elements. Common methods include feature jittering \cite{margeloiu2024tabmda}, cell completion \cite{zhang2019auto}, table decomposition and reconstruction \cite{fang2022semi}, and random row-wise permutation. Feature jittering techniques generally aim to learn embeddings of the target tabular data and perform suitable manipulations in the embedding space.  Cell completion involves populating empty cells, supplying additional entities or missing attribute in the target tables.  With decomposition and reconstruction techniques, entire tables are disentangled, simplified and recomposed into new tables. The relative positions of rows can also be altered in various ways to provide additional variation of the tabular data.  
 
 %%%%%%%%%%%%%%%% FIGURE 4   %%%%%%%%%%%%%%%%%%%%%%%%%%%%%%
 %\usepackage{float}
 \begin{figure*}[!htb]
 	\vspace {0mm}
 	\centering
 	\includegraphics[width=0.9 \linewidth]{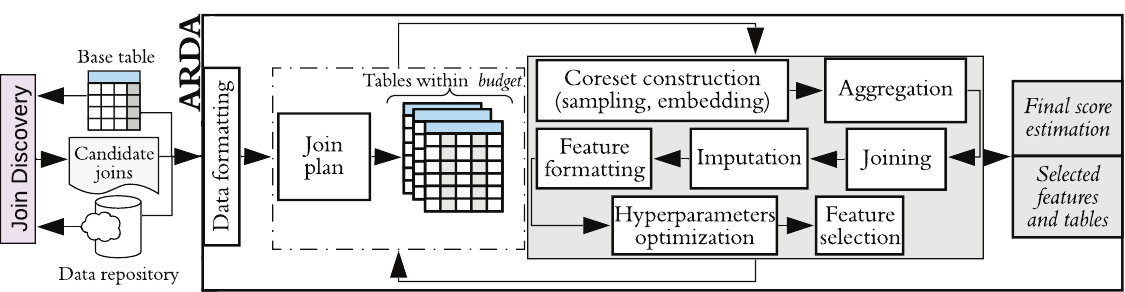} \vspace {0mm}
 	\caption{ARDA \cite{chepurko2020arda}, an example of a data acquisition technique that relies on integrating  data from multiple sources. In addition to performing feature engineering operations, the technique searches for optimal hyperparameters to integrate data from related tables. 
 	}\label{fig:6_Data_integration}
 \end{figure*}
 %%%%%%%%%%%%%%%%% End of Figure 5 %%%%%%%%%%%%%%%%%%%%5
 
 \subsection{Data integration}
 
 Another common approach to data augmentation is to combine several complementary datasets in order to obtain richer and expanded training data. The first step in the integration process is data discovery, where relevant datasets per the given machine learning problem and objective are identified and designated as candidate datasets. The second step of the process consists in selecting suitable data points from the discovered datasets for use. This involves collecting data or extracting relevant data elements from different sources, applying necessary transformations and additional processing to make them homogeneous and compatible. The third step is to then reconstitute them into a new target dataset. This process is shown in Figure \ref{fig:6_Data_integration}. 
 
 \subsubsection{Image data integration}
 
 Research (e.g.,  \cite{bazrafkan2017deep,gao2021searching,gao2023meta}) has shown the effectiveness of mixing different image datasets to overcome the limitations of smaller datasets. Despite the attractiveness of this method, it is particularly challenging to integrate image data from different sources as these data often suffer from problems of inconsistent, incomplete, or noisy (i.e., inaccurate) annotations, rendering the effectiveness of automated data integration methods for image modality low. Moreover, data cleaning methods are ill-suited for image modalities. Because of these difficulties, only a few works have attempted using automated machine learning methods to integrate image data from varied sources. Some researchers (e.g., Kim et al.  \cite{ kim2021lada, yao2020searching}) have proposed to address this challenge by employing dedicated label correction or filtering algorithms in the integration framework. 
 
 Instead of attempting to clean noisy image data or to correct erroneous labels, an alternative class of approaches \cite{gao2021searching,gao2023meta,shu2023dac,gao2024dense,gao2022loss} aims to modify the training process in such a way that makes the resulting model more robust to the impact of noisy image labels. For instance, Gao et al.  \cite{gao2021searching} propose a so-called Automated Robust Loss (ARL) framework, an AutoML-based meta-learning method, that learns loss functions that are robust to noisy labels. Yao et al. \cite{yao2020searching} employ a function approximation method based on a modified Newton algorithm within an AutoML framework to filter out inconsistent and noisy labels by selectively sampling only images with clean labels. 
 
 \subsubsection{Tabular data integration}
 
 Techniques for the integration of tabular data \cite{suri2021ember,chepurko2020arda,bai2021atj,li2021data,kumar2015learning,esmailoghli2021cocoa,koutras2021valentine,li2021ai} generally aim to automatically discover, select and aggregate related data in order to extend a given dataset. Tabular data is typically characterized by problems such as missing values, large discrepancy in data representation, inconsistency of keys, and the presence of a wide variety of variables. This makes the problem of integrating this type of data an extremely challenging but useful task. Chepurko et al. \cite{chepurko2020arda} presented an AutoML-based data acquisition strategy aimed at integrating different but related tabular data into a single dataset to provide effective training of machine learning models. Given a set of database tables and corresponding keys, the proposed model automatically searches for, and joins the most related tables in an optimal way. Bai et al. \cite{bai2021atj} propose neural architecture search (NAS) \cite{elsken2019neural} method to find the best network topology and connections that provide the most effective integration.   Kumar et al. \cite{kumar2016join} demonstrate that integrating relational data from multiple sources does not always lead to performance improvements; it requires a rigorous and careful implementation to achieve improvements.   It is the task of the AutoML framework to find integration strategies that provide optimal performance. 
 
 \subsubsection{Integrating text and multimodal data}
 Approaches have been devised to allow automated machine learning models to seamlessly integrate text \cite{mustafa2021automated,nikitin2022automated} as well as heterogeneous and multimodal data \cite{shi2021multimodal,erickson2022multimodal}. Owing to the differences in construction of the different datasets, it often requires enormous preprocessing to achieve good integration.   Automated machine learning techniques are able to handle all the needed processing tasks for integrating multi-modal data.  For instance, Shi et al. \cite{shi2021multimodal} propose to employ specialized transformer models as multimodal data processing units to allow automated learning on multimodal data – specifically, text and tabular information in different formats. The Transformer units rely on natural language processing (NLP) to obtain useful features from text datasets. In the processing stage, these feature sets are then aggregated, transformed and combined with tabular data, where AutoGluon-Tabular \cite{erickson2020autogluon}, an AutoML framework for learning on tabular data, is then used to seamlessly processed the aggregated information from multiple data sources and formats. 
 
 Another challenge is that data from different sources may exhibit different statistical distributions \cite{nargesian2022responsible,chai2022selective}. Moreover, in situations where the collected data is from sources created by diverse players with different levels of expertise or conflicting interests and goals, data quality and bias become important issues and need to be handled in the integration process \cite{nargesian2022responsible}. This is currently achieved using automated  cleaning methods \cite{shende2022cleants}.

 %%%%%%%%%%%%%%%% FIGURE 5   %%%%%%%%%%%%%%%%%%%%%%%%%%%%%%
 %\usepackage{float}
 \begin{figure*}[!htb]
 	\vspace {0mm}
 	\centering
 	\includegraphics[width=0.8 \linewidth]{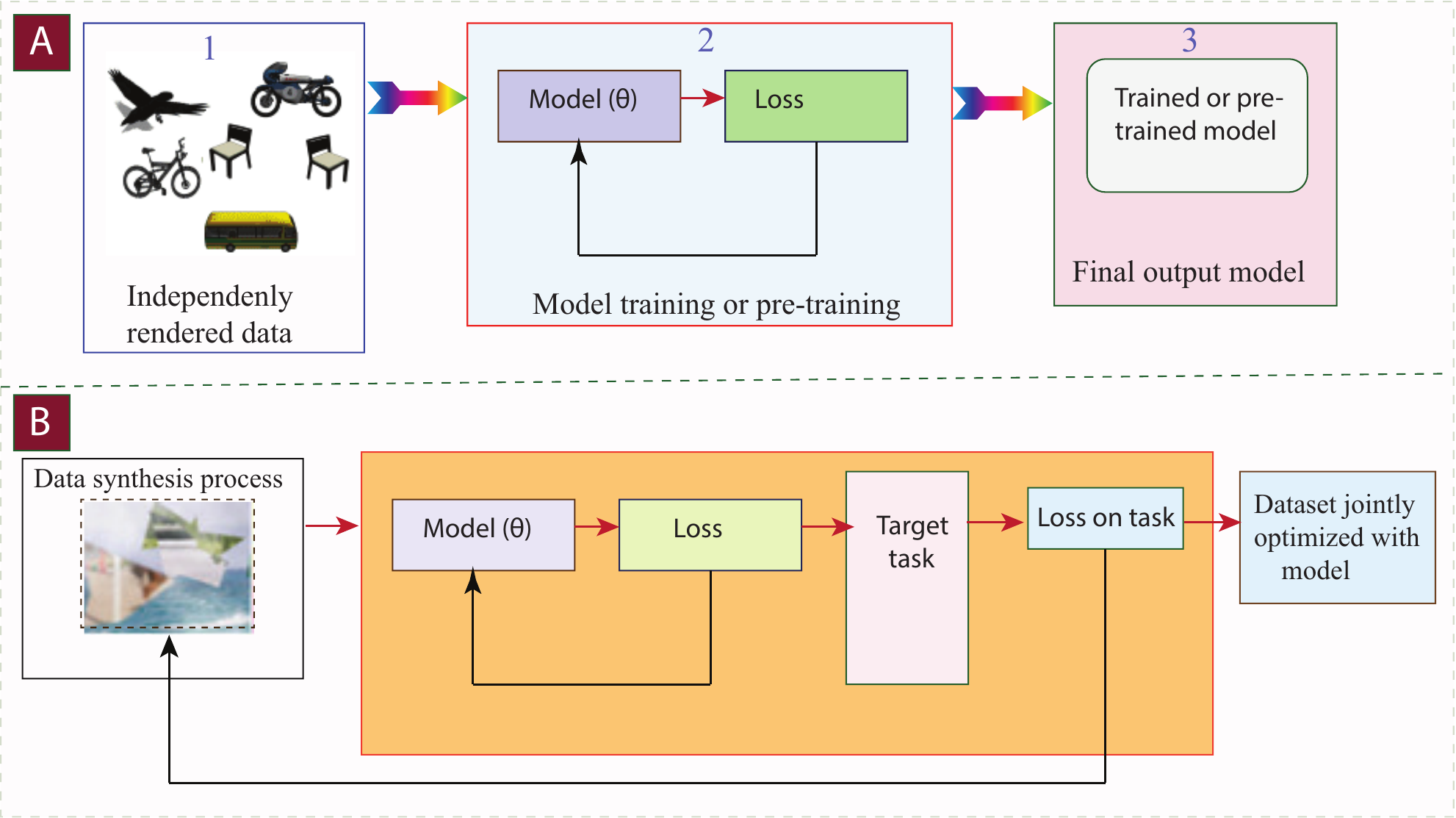} \vspace {0mm}
 	\caption{A comparison of classical and automated data synthesis approaches. Classical methods of data synthesis involve three distinct steps (A): (1) creation of primitive data elements (e.g., geometric models), (2) model training, and (3) evaluation of results and fine-tuning.  With AutoML approaches,however, th entire data synthesis process is carried out end-to-end in a single process. Like in many AutoML procedures, the process typically utilizes a bi-lel optimization scheme to optimize hyperparameters for both synthesis primitives (outer loop)  and model hyperparameters (inner loop).
 	}\label{fig:7_DataSynthesis_vs_Trad_DA}
 \end{figure*}
 %%%%%%%%%%%%%%%%% End of Figure 5 %%%%%%%%%%%%%%%%%%%%5
 
 \subsection{Data synthesis with AutoML}

 Data synthesis methods effectively generate new data samples from “scratch”. The use of synthetic data \cite{mangrulkar2020image,nikolenko2021synthetic,de2021next,tremblay2018training}  for training machine learning models has emerged as an important approach to address data scarcity issues in many domains. The generation of synthetic data is currently a tedious and time consuming process. Data synthesis for medical image analysis \cite{mangrulkar2020image,starly2005three} and other computer vision tasks \cite{chang2015shapenet}, for example, typically involve mundane processes of creating and manually manipulating representation primitives such as basic geometric shapes and simulation parameters to produce realistic data.
 
\begin{figure*}[!htb]
 	\vspace {0mm}
 	\centering
 	\includegraphics[width=0.75 \linewidth]{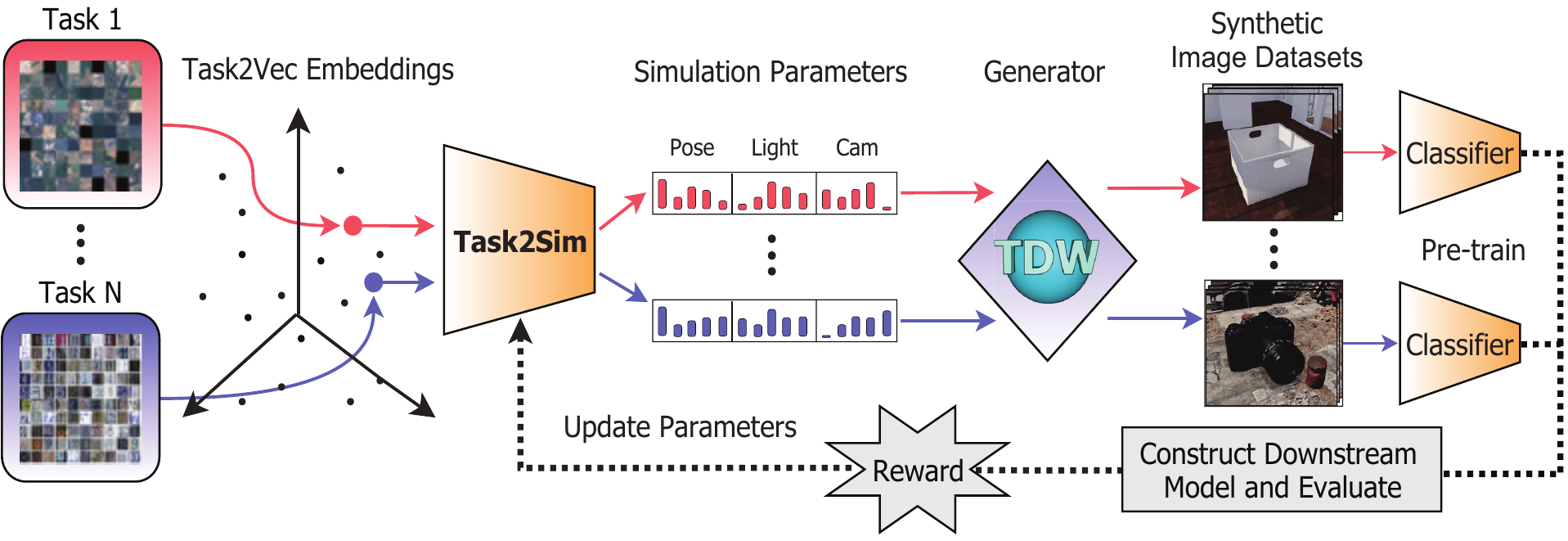} \vspace {0mm}
 	\caption{Task2Sim \cite{mishra2022task2sim} generates synthetic data from primitive elements and utilizes reinforcement learning-based gradient approximation method known as REINFORCE algorithm \cite{williams1992simple} to estimate the gradients of the performance on the downstream task with respect to the data simulation and model parameters. The model first learns to map input images to the best set of simulation primitives. It is then trained to generate synthetic image samples for specified tasks in a pretraining phase. The aim is to transfer this ability to unseen tasks.
 	}\label{fig:8_Task2VecA}
 \end{figure*}

\begin{figure*}[!htb]
 	\vspace {0mm}
 	\centering
 	\includegraphics[width=0.75 \linewidth]{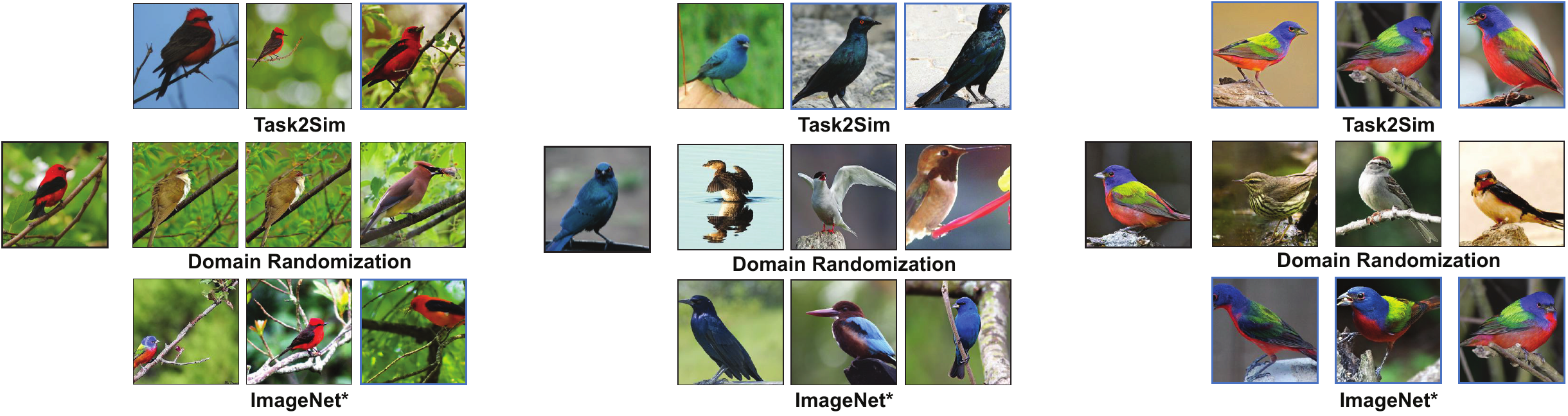} \vspace {0mm}
 	\caption{Sample synthetic image data generated by Task2Sim model proposed by Mishra et al. \cite{mishra2022task2sim}—top images in each column. After pretraining, the model is able to generate new samples for unseen tasks. Here, given an exemplar image, different views are generated by pretrained Task2Sim model and two other methods, domain randomization and ImageNet pretraining.
 	}\label{fig:9Task2Vec_B}
 \end{figure*}

 Unlike classical data acquisition methods that perform synthesis separately as a distinct subproblem, approaches based on automated machine learning principles typically treat the synthesis task and  the target application as a unified problem, where optimal simulation parameters can be jointly learned for the downstream task. This distinction is shown in Figure \ref{fig:7_DataSynthesis_vs_Trad_DA}. The workflow of a typical data synthesis method, Task2Sim \cite{mishra2022task2sim}, is shown in Figure \ref{fig:8_Task2VecA}. Figure \ref{fig:9Task2Vec_B} shows samples of image data generated by the technique. The approach contrasts sharply with traditional approaches such as A3D \cite{wang2022a3d} that do not make use of feedback signals from the performance on the downstream task to optimize simulation parameters. The synthesis process in the latter case, thus, involves extensive manual tuning.     
 
 \subsubsection{Image data synthesis}
 
 AutoML approaches aimed at automating synthetic image data generation pipelines include \cite{du2021auto,kim2021lada,thompson2022augmenting,owoyele2022application}. For example, Kim et al. \cite{kim2021lada} propose a deep neural network that aims to collect real world data by means of adaptive sampling. Their model, LADA, is both a data integration and synthesis method as it learns to acquire informative samples and at the same time generates synthetic instances from the acquired samples using AutoML-based data synthesis principles. In order to generate new samples, they train a policy to maximize the acquisition score based on feedback of the training loss from a classification network.  To guarantee the informativeness of the synthetic samples, the authors propose a so-called look-ahead data acquisition technique that speculates about (i.e., predicts) the quality of plausible candidate data in advance. Also, they employ an \textit{oracle} to annotate the unlabeled real-world data acquired.  
 
 %

 %%%%%%%%%%%%%%%% FIGURE 6   %%%%%%%%%%%%%%%%%%%%%%%%%%%%%%
 %\usepackage{float}
 %%%%%%%%%%%%%%%%% End of Figure 8 %%%%%%%%%%%%%%%%%%%%5

 %
 %%%%%%%%%%%%%%%% FIGURE 7   %%%%%%%%%%%%%%%%%%%%%%%%%%%%%%
 %\usepackage{float}
 %%%%%%%%%%%%%%%%% End of Figure 9 %%%%%%%%%%%%%%%%%%%%5
 
\begin{figure*}[!htb]
 	\vspace {0mm}
 	\centering
 	\includegraphics[width=1.0 \linewidth]{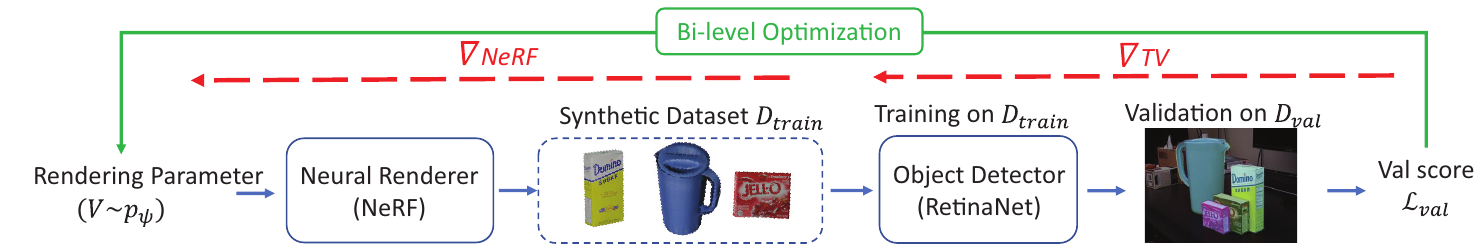} \vspace {0mm}
 	\caption{Basic architecture of NeuralSim model \cite{ge2022neural}. The approach is a typical AutoML-based data synthesis models that employ a bi-level optimization scheme to fintune model hypeparameters with the help of multiple feedback signals. The outer loop optimizes hyperparameters for rendering primitives while the inner loop the machine learning parameters.
 	}\label{fig:11_NeuralSim_DA}
 \end{figure*}

 In computer vision domains (e.g., \cite{behl2020autosimulate,ruiz2018learning,shirobokov2020black,sun2021autoflow,mishra2022task2sim,kar2019meta}), the task is usually to synthesize visual scenes or images. The search space in this case consists of mathematical operations describing geometry primitives and other scene parameters such as textures, color, lighting, pose and camera properties. The synthesis process involves the conversion of these elementary information into realistic 2D or 3D data. The role of automation in the synthesis process is to jointly optimize the rendering parameters and the machine learning model conditioned on the performance of the synthetic data on the target task. Ruiz et al. \cite{ruiz2018learning} utilize AutoML technique with reinforcement learning-based optimization to control the quality of synthetic data generated using computer graphics modeling method. Their approach aims to automatically find scene parameters for the simulation process that maximize the accuracy of trained model. More modern techniques such as NeuralSim \cite{ge2022neural} (Figure \ref{fig:11_NeuralSim_DA}) employ neural radiance fields as implicit representations of simulation primitives instead of relying on explicitly modeled geometric primitives. The method has been designed to synthesize data for object detection tasks. It employs a bi-level optimization strategy to jointly optimize model parameters of the object detector and rendering parameters of the synthetic data (e.g., illumination, texture, object pose, material, color and appearance).

 %%%%%%%%%%%%%%%% FIGURE 8   %%%%%%%%%%%%%%%%%%%%%%%%%%%%%%
 %\usepackage{float}
 %%%%%%%%%%%%%%%%% End of Figure 12 %%%%%%%%%%%%%%%%%%%%

 Another area where automated data synthesis has been particularly successful is in tasks such as depth estimation, optical flow and stereo vision. Since these tasks do not require photorealistic data, the synthesis process is less resource-intensive from the computational perspective. Consequently, models pre-trained on synthetic datasets consistently outperformed those trained on real data alone. Because of this success, techniques to automate the synthesis of data \cite{sun2021autoflow,han2022realflow,kar2019meta}  in these domains are currently receiving serious  attention.  Sun et al. \cite{sun2021autoflow}, for example, propose a technique, called AutoFlow,  to automatically render synthetic 2D data for optical flow tasks by learning model hyperparameters to optimize the appearance, shape and motion of the generated data. The synthetic data generated by AutoFlow is designed to be used to pre-train optical flow models before fine-tuning on real data. Unlike classical approaches (e.g., \cite{kortylewski2019analyzing,mishra2022task2sim,ebadi2022psp}) that tackle the pre-training task as an independent process, Sun et al. \cite{sun2021autoflow} framed the data synthesis and pre-training performance (i.e., performance on target data) as a joint optimization problem.  They demonstrate that the approach achieves better performance than conventionally rendered synthetic 3D datasets such as FlyingThings3D \cite{mayer2016large} and Flying Chairs \cite{dosovitskiy2015flownet}. Moreover, their approach has proven to be significantly more data-efficient in pre-training optical flow models on MPI-Sintel \cite{butler2012naturalistic} and KITTI \cite{geiger2012we} datasets than traditional approaches. 
 
 \subsubsection{Tabular data synthesis}
 Some of the most popular traditional methods for generating synthetic tabular data involve random oversampling. This approach is particularly useful for balancing imbalanced data, where minority classes are typically sampled and interpolated to create new datapoints. SMOTE \cite{chawla2002smote}, a seminal work among this class of approaches, is commonly used to generate synthetic data to populate tables by means of k-Nearest Neighbor algorithm which interpolates between existing data points to create new samples. These oversampling methods can be incorporated within larger neural frameworks to automate the data synthesis process . For instance, Wang and Pai \cite{wang2023enhancing} propose an approach that employs SMOTE together with a GAN model to generate novel tabular clinical data. Similarly, DeepSMOTE \cite{dablain2022deepsmote} combines SMOTE with an encoder-decoder sub-model for synthetic data generation. Aragao et al.  \cite{aragao2024enhancing} propose a self-balancing, synthetic tabular data generation pipeline that utilizes both undersampling and oversampling techniques within an AutoML framework to generate new data. Some new approaches bypass the step of modeling data analytically and instead learn the generation process end-to-end from data. Rashidi et al. \cite{rashidi2023stng}, for example, devise a so-called Synthetic Tabular Neural Generator (STNG) that trains a dedicated neural network within a larger AutoML framework to  specifically generate rich tabular data.

 \subsubsection{Text data synthesis}
 In the past, models struggled to generate meaningful text automatically. Approaches often utilize shemantic parsers together with dedicated templates to simplify the process. AutoQA \cite{xu2020autoqa}, for example, employs an end-to-end neural paraphrasing model to generate answers (i.e., text) to different questions using template-based parsing technique. Large language models have recently revolutionized text generation in NLP domains \cite{li2024data,glazkova2024evaluating}. Consequently, they have been increasingly used to generate text for AutoML pipelines \cite{xu2024large}. Large language models are extremely powerful and flexible for this use case. They are adept at producing semantically meaningful text on many different contexts and tasks. Generally, in this process, the LLM framework serves asan interface for not only text generation but also for performing additional functions, including model configuration (see \cite{tornede2023automl,xu2024large}). 
 mposed in many different ways. We discuss the most common classes of techniques next. These are grouped into three main approaches: (1) methods based on fixed or predefined data augmentation operations, (2) deeply learned data transformation operations, and (3) methods based on neural architecture search \cite{ma2023pareto,ma2024single,sun2020automatically} – i.e., learning the best neural architectures for the given data.

 \section{Composition of search space} \label{section:SearchComposition}
 
 The search space can be composed in many different ways. We discuss the most common classes of techniques next. These are grouped into three main approaches: (1) methods based on manual or pre-defined data augmentation operations, (2) deeply learned data transformation operations, and (3) those based on neural architecture search (NAS) – i.e., learning the best neural architectures for the given data.  

\subsection{Fixed augmentation operations}

Many AutoML techniques (e.g., \cite{cubuk2019autoaugment,cubukpractical,liu2021direct,hataya2020faster,lim2019fast, lin2021local,li2020differentiable,lu2023bda}) employ a fixed set of pre-defined transformation operations to perform data augmnetation. In these works, the transformation operations themselves are fixed and can only be varied by learning tunable parameters that control hyperparameters such as augmentation intensities and application probabilities. Thus, the task of the AutoML process in this case is reduced to simply learning the most useful combination of these fixed augmentations along with the optimal settings for these hyperparameters.  In the procedure, the choice of specific transformation operations and their corresponding magnitudes to include in the search space is based on domain knowledge and intuition. Fig. 9 shows the common methods typically employed by approaches that utilize pre-defined transformation operations to perform data augmentation.

%%%%%%%%%%%%%%%% FIGURE 8: Search space condruction %%%%%%%%%%%%%%%%%%%%%%
%\usepackage{float}
\begin{figure}[H]
	\vspace {-3mm}
	\centering
	\includegraphics[width=0.95 \linewidth]{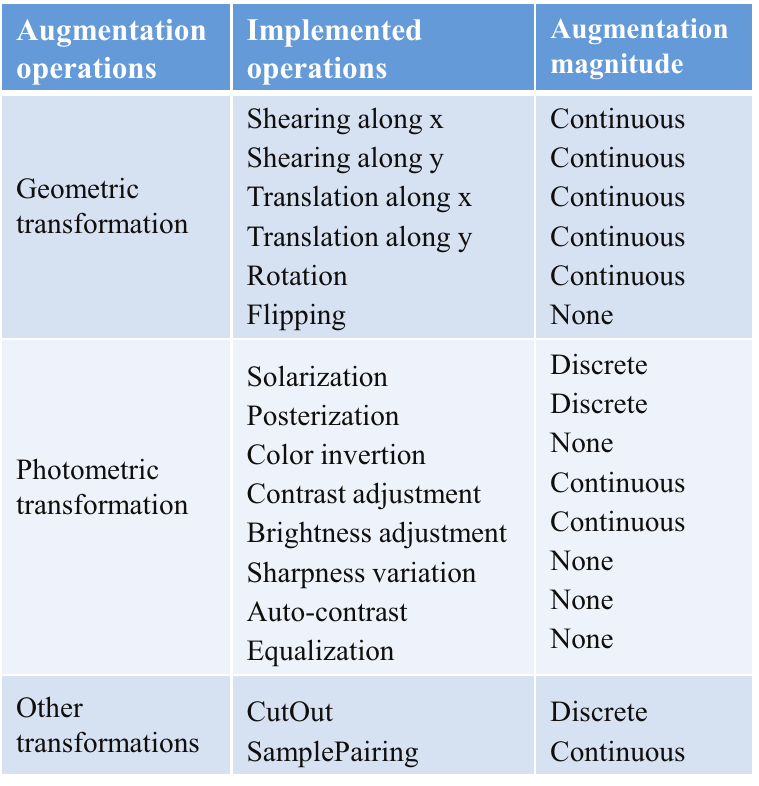} \vspace {-3mm}
	\caption{Composition of fixed augmetation operations. Many works–including AA \cite{cubuk2019autoaugment}, FastAA \cite{lim2019fast}, FasterAA \cite{hataya2020faster}, PAA \cite{lin2021local}, DADA \cite{li2020differentiable} and BDA \cite{lu2023bda}–employ this or similar search space construction.
	}\label{fig:10BiLevelOptimization}
\end{figure}
%%%%%%%%%%%%%%%%% End of Figure 8 %%%%%%%%%%%%%%%%%%%%

The process of composing predefined augmentation operations is typically sub-divided into three main sub-problems: (1) composition of transformation operations to be used for augmentation, (2) a specification of (mostly) discrete probabilities with which the transformation operations would be applied on the input data, and (3) a specification of the applicable transformation magnitudes (i.e., augmentation intensities). Again, for simplicity, transformation magnitudes are often discrete values. 
The original work, AutoAugment (AA) \cite{cubuk2019autoaugment}, for example, uses a set of 16 common image transformation operations to augment image data for classification tasks. It trains a deep CNN model on each set of transformation operations to learn optimal parameters for effective image augmentations using validation data. Many subsequent works (e.g., \cite{li2020differentiable,hataya2020faster,lim2019fast,kashima2020joint,zhang2019adversarial}) adopt a similar formulation of augmentation operations in the search space as in AA \cite{cubuk2019autoaugment} but utilize more efficient optimization methods and additional mechanisms to reduce computational requirements. For instance, AdvAA \cite{zhang2019adversarial} maintained the search space specification in AA \cite{cubuk2019autoaugment} but, instead of optimizing the augmentation policy hyperparameters and network parameters separately, jointly optimizes both tasks through adversarial training. Also, instead of RL, a gradient approximation method is used for the policy search. The visual results of applying fixed transformation operations with different magnitudes on image data are shown in Fig. 10. The sample results indicated are specifically for augmentation policies of FasterAA \cite{hataya2020faster}.

%%%%%%%%%%%%%%%% FIGURE 9: n %%%%%%%%%%%%%%%%%%%%%%
%\usepackage{float}
\begin{figure}[H]
	\vspace {-3mm}
	\centering
	\includegraphics[width=1.0 \linewidth]{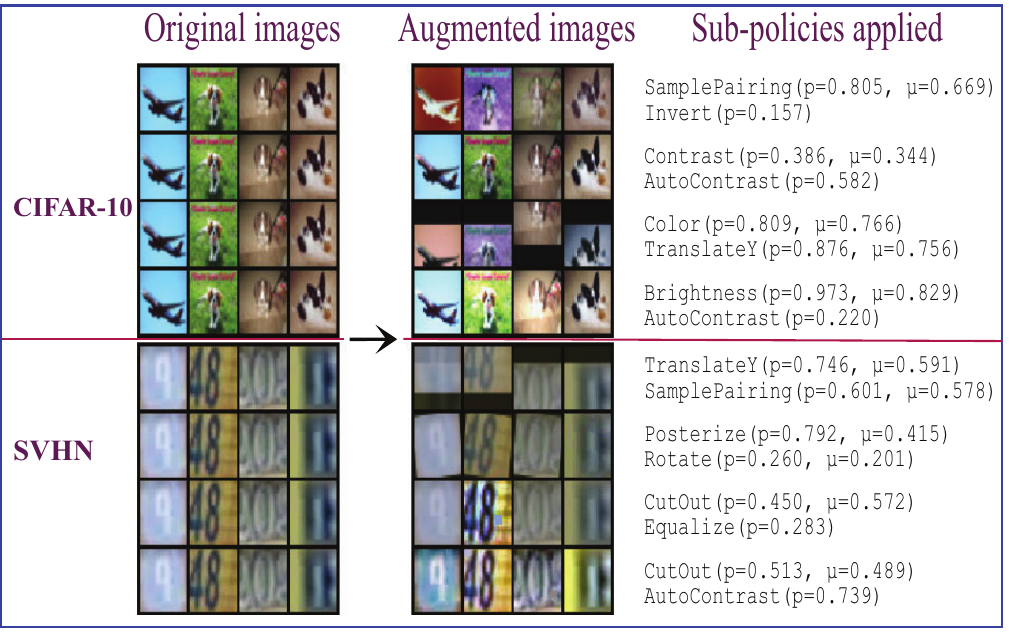} \vspace {-3mm}
	\caption{Augmentation policies of FasterAA \cite{hataya2020faster} and the result of applying these policies on sample images of the  CIFAR-10 \cite{krizhevsky2009learning} and SVHN \cite{netzer2011reading} datasets.
	}\label{fig:9AugPoliciesFastAA}
\end{figure}
%%%%%%%%%%%%%%%%% End of Figure 9 %%%%%%%%%%%%%%%%%%%%

While approaches such as AutoAugment \cite{cubuk2019autoaugment}, FastAA \cite{lim2019fast}, FasterAA \cite{hataya2020faster} and \cite{wei2020circumventing} associate magnitudes to each operation as transformation-dependent attributes, Direct Differentiable Augmentation Search (DDAS) \cite{liu2021direct} treats different transformation magnitudes as unique augmentations. This representation simplifies the search space and greatly reduces the computational demand for performing subsequent search operations to optimize model parameters. Many recent approaches \cite{cubukpractical,lin2019online} follow this philosophy of designing more simplified search space that facilitate efficient optimization. 

The task of creating effective augmentation policies based on manually-designed, fixed transformation operations has a number of significant drawbacks. First, the process is tedious, time-consuming and is highly dependent on domain expertise. Second, the approach constrains the degree of automation that can be achieved. More significantly, given the large space of possible augmentations, it is not practical to manually incorporate all potentially useful augmentation operations, thereby further limiting the benefits that can be derived from automation. In view of the limitations of composing a search space consisting solely of fixed, manually-designed transformation operations, some new approaches have proposed learning more useful augmentations from data.

\subsection{Learned augmentation operations}

As already noted, transformation operations used in most current automated data augmentation pipelines are designed manually. This constrains their scope to the range of intuitive and discrete transformation routines that can be constructed analytically. To further extend the capabilities of automated machine learning-based data augmentation methods, more recent works \cite{mounsaveng2021learning,gao2021enabling,zhao2019data,miaomulti} attempt to automatically learn useful augmentation operations directly from the training data. The design of these types of operations requires less domain expertise to compose task-specific augmentation policies. These approaches have already demonstrated very competitive results. In many cases, the performance of models designed using learned augmentations are superior to those employing fixed augmentation operations.

\subsubsection{Fully learned augmentations}
An alternative to learning to dynamically modify primitive transformation operations is to deeply learn the transformation functions from data. Techniques such as spatial transformer networks (STN) \cite{jaderberg2015spatial}, Deformable Convolutional Networks (DCN) \cite{dai2017deformable} and \cite{zhao2019data} have demonstrated the effectiveness of learning effective data transformation operations using deep learning models.  Zhao et al. \cite{zhao2019data} formulate the data augmentation problem as a composite transformation function consisting of spatial transformation defined by voxel displacement and an appearance transformation defined by pixel intensity transformations. Chu  et al. \cite{chu2022augmentation} propose a natural language processing (NLP) model consisting of three sub-models—an encoder, a custom augmentation network and a document classification model—and then find the best augmentation strategies by employing a reinforcement learning technique to jointly optimize all sub-models. 

The STN, in particular, has recently been widely used within automated machine learning models (e.g., \cite{mounsaveng2021learning,miaomulti,tang2020onlineaugment,kim2021lada}) as a means of generating the basic transformation operations that can be used to form data augmentation policies. For example, Mounsaveng et al. \cite{mounsaveng2021learning} employ STN-based augmenter networks to learn effective transformations from data and use a bi-level optimization scheme to search for model and augmentation parameters. Similarly, Miao and Rahman \cite{miaomulti} incorporate a Spatial Transformer in a network structure based on AutoAugment \cite{cubuk2018autoaugment} to learn generic affine transformations in a traffic sign classification task. Kim et al. \cite{kim2021lada} employ an STN-based model known as InfoSTN together with InfoMixup, an adaptive form of MixUp \cite{zhang2017mixup}, for automated data acquisition and further augmentation.

%%%%%%%%%%%%%%%% FIGURE 10+1: Search space condruction %%%%%%%%%%%%%%%%%%%%%%
%\usepackage{float}
\begin{figure}[H]
	\vspace {0mm}
	\centering
	\includegraphics[width=1.0 \linewidth]{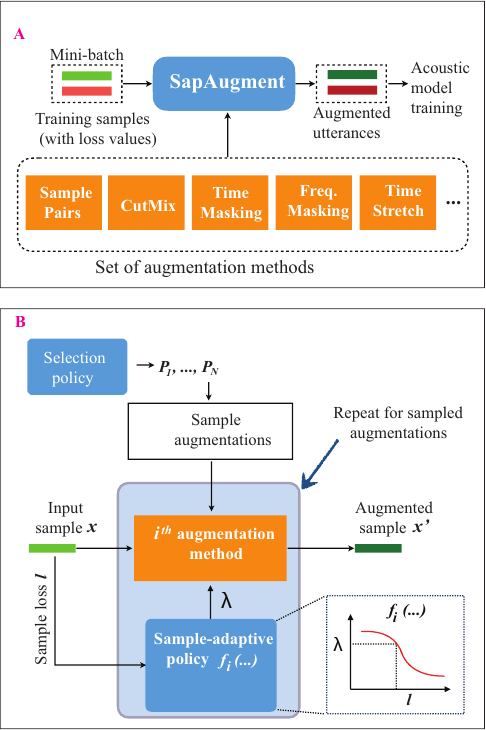} \vspace {0mm}
	\caption{Sample-Adaptive Policy for Augmentation (SapAugment) \cite{hu2021sapaugment} uses basic transformation operations and simple data augmentation methods such as SamplePairing \cite{inoue2018data} and CutMix \cite{yun2019cutmix} to transformation input data (A). The approach additionall implement a mechanism to adapt the learnable parameters of these basic operations to the specific input dataset according to the training loss (B). 
	}\label{fig:11SapAugment}
\end{figure}
%%%%%%%%%%%%%%%%% End of Figure 8 %%%%%%%%%%%%%%%%%%%% 

A large number of learned augmentation methods (e.g., \cite{gao2021enabling,zhang2019adversarial,chinbat2022ga3n,tang2020onlineaugment,zhao2019data,peng2018jointly}) utilize generative modeling techniques to automatically generate augmentations instead of relying on manual image transformation operations. Techniques based on generative adversarial networks and variational autoencoders are particularly useful in this regard. They are able to synthesize realistic data with rich appearance variations. Augmentation policies are then learned to select the most effective augmentations based on training and evaluating on the large synthetic set generated. For instance, Gao et al. \cite{gao2021enabling} propose a fully automated data augmentation pipeline that employs three separate GAN sub-models to perform different image transformations in order to generate augmented data– an appearance perturbation sub-model, as well as global and local affine transformation units.  They then use a game theoretic formulation based on two min-max \cite{liu2019min} optimization scheme to learn effective augmentations.   Similarly, Chinbat and Bae \cite{chinbat2022ga3n} constituted a learned search space of synthetic data using a GAN model as a policy network. Another interesting implementation of learned augmentations based on generative modeling can be found in works such as \cite{tang2020onlineaugment}. Instead of directly specifying the augmentation operations, a complex deep learning model was used to learn appropriate transformations from data. Here, the authors employed a so-called Augmentation STN (A-STN) that was a modification of the original STN model, proposed by Jaderberg et al. \cite{jaderberg2015spatial}, to learn affine transformations. They also incorporated a complementary model based on variational Autoencoder architecture [20]– known as Deformation VAE (or D-VAE) –that performs local deformations on input samples or feature maps. A third model, known as perturbation VAE (or P-VAE) was used to learn photometric augmentations such as brightness and contrast adjustment, color jittering and noise addition.  These three augmentation networks were jointly trained with the target task to learn effective augmentation strategies. 

One of the most important advantages of methods based on deeply learned transformation operations is their universality – ability to generalize well in a large variety of scenarios without prior knowledge about the target tasks.  Yet in some practical situations, it is more effective and beneficial to leverage domain knowledge to encode important attributes for specific tasks \cite{lee2020human,li2021automl}.

\subsubsection{Learning dynamic augmentations from basic transformation operations}

An alternative to automatically learning data transformation operations directly from data (e.g., see \cite{chinbat2022ga3n,tang2020onlineaugment,zhao2019data}) is to compose simple and flexible transformation functions whose properties can be modified in the process of training.  In contrast to approaches such as AA \cite{cubuk2019autoaugment}, FastAA \cite{lim2019fast} and FasterAA \cite{hataya2020faster} that compose a search space consisting of fixed augmentations and evaluate different combinations of these to obtain effective policies, many recent approaches (e.g., \cite{zheng2022deep,hu2021sapaugment,liu2021divaug,ho2019population}) propose to learn augmentation policies in a more flexible and dynamic way. They design the search space as a composition of loosely-specified primitive data transformation operations that can be modified in the course of training to generate better augmentations. This flexibility allows the model to adapt the augmentation policies to the specific task, dataset and, in some cases (e.g., see \cite{ho2019population}), specific input instances.  Population Based Augmentation (PBA) \cite{ho2019population}, for example, utilizes an evolutionary algorithm technique that allows augmentation policies to generate child policies which evolve over time in the course of training. Also, in SaPAugment \cite{hu2021sapaugment}, Hu et al. propose a so-called sample-adaptive policy that dynamically adapts transformation parameters according to the training loss (see Figure \ref{fig:11SapAugment}). Even though the search space (i.e., the set of primitive augmentation operations) is fixed like in AutoAugment \cite{cubuk2019autoaugment}, FastAA \cite{lim2019fast} and FasterAA \cite{hataya2020faster}, SapAugment (SAPA) \cite{hu2021sapaugment} is able to adaptively learn and apply different transformation parameters to different datasets in a context dependent manner. Chen et al. \cite{chen2022scale} proposed an automated augmentation method for object detection tasks based on learning dynamic augmentations that are robust to scale variations. Deep AutoAugment (DeepAA) \cite{zheng2022deep} eliminates the need for manually constructed default augmentation policies in the search space by replacing them with sequentially stacked augmentation layers which can be trained to generate more nuanced augmentation policies. Adversarial AutoAugment (AdvAA) \cite{zhang2019adversarial} is aimed at generating dynamic augmentation policies through adversarial training.

\subsubsection{Instance-adaptive dynamic search space}

Instead of employing a single augmentation strategy, many recent works such as MetaAugment \cite{zhou2021metaA}, AdaAug \cite{cheung2021adaaug}, InstaAug \cite{miao2022learning} and \cite{chu2022augmentation} have proposed learning adaptive, instant-dependent data augmentation policies from data. These methods are based on the observation that certain transformations may only produce useful augmentations for specific input types but can be harmful when applied globally. For example, whereas the interpretation of the digit "8" is unaffected by flippng, a vertical flipping of the latin symbol “R” may change it to the 
\begin{otherlanguage*}{russian}
	Cyrillic letter “Я”,
\end{otherlanguage*}
thereby corrupting its semantic interpretation. Similarly, a 180$^{ o}$ rotation of the Arabic numeral “6” may change its semantic label to “9” whereas the letter "O" is generally agnostic to such a transformation. Therefore, instead of applying the same set of transformation operations on all input samples, instance-adaptive methods \cite{zhou2021metaA,cheung2021adaaug,miao2022learning,chu2022augmentation} learn  input specific transformations according to the dataset and task. 
 
 These approaches focus on realizing more fine-grained transformation parameterizations that provide flexible augmentations which can be applied to individual instances instead of using coarse transformations that satisfy all samples of the entire dataset. In this formulation, the instance-level augmentations correspond to specific configurations of transformation hyperparameters. MetaAugment \cite{zhou2021metaA} utilizes an auxiliary model within an AutoML framework which re-weights input samples in order to learn the application probabilities of different transformation operations for specific input instances. Thus, by learning the probability of applying transformations for different instances, the sub-model essentially predicts useful augmentations that may be specific to the underlying instance. Based on a similar philosophy, the developers of AdaAug \cite{cheung2021adaaug} propose an adaptive data augmentation technique that, like \cite{yoo2023class}, learns effective augmentation policies in a category (e.g., \cite{yoo2023class})- and sample (e.g., \cite{cheung2021adaaug})-dependent way. Their model re-uses extracted features from the input data to map samples to useful transformations, thus adapting the augmentation operations for each instance. The technique is specifically aimed at enhancing the generalization ability of models trained with this augmentation method.  The authors claim that data augmentation policies learned by AdaAug can perform well on new datasets without further fine-tuning.   Using a slightly different approach, InstaAug \cite{miao2022learning} learns instance-adaptive augmentation strategies by mapping instances to distributions of transformation operations instead of single transformations, which are relatively coarse. The authors \cite{miao2022learning} propose to map input samples to transformation distribution parameters in order to learn instance-specific data augmentations policies from training data. Similar to MetaAugment, InstaAug \cite{miao2022learning} incorporates a dedicated neural network sub-model, so-called augmentation module, to perform the mapping of input instances to desired transformations. In the training process, the augmentation module samples transformations from the set of transformation distributions and applies them to individual input instances in order to generate augmented samples. The resulting augmented samples are fed into a classification network which is then trained to make predictions and, hence, provides information about the quality of the augmentations. A  comparison of the main search approaches is presented in Table \ref{tab:my-table1}.

\begin{table*}[h!]

	\caption{A comparison of the main search space approaches for automated data augmentation}
	\label{tab:my-table1}
	
		\scalebox{0.80}{
			
	\begin{tabular}{@{}llll@{}}
		\toprule
		\textbf{Main direction} &
		\textbf{Example works} &
		\textbf{Main strengths} &
		\textbf{Weaknesses} \\ \midrule
		\begin{tabular}[c]{@{}l@{}}Fixed\\ augmentations\end{tabular} &
		\begin{tabular}[c]{@{}l@{}}AA  \cite{cubuk2019autoaugment}, DDAS   \cite{liu2021direct}, \\    \\ \cite{lim2019fast},FasterAA   \cite{hataya2020faster}, PAA \cite{lin2021local}\end{tabular} &
		\begin{tabular}[c]{@{}l@{}}Easy to design and implement\\    \\ Intuitive and transparent \\    \\ Decreased search space\end{tabular} &
		\begin{tabular}[c]{@{}l@{}}Limited scope   of data augmentation operations\\    \\ Requires   domain expertise\end{tabular} \\
		\begin{tabular}[c]{@{}l@{}}Learned\\ augmentations\end{tabular} &
		\begin{tabular}[c]{@{}l@{}}GA3N   \cite{chinbat2022ga3n}, AdvAA \cite{zhang2019adversarial},\\ OnlineAugment   \cite{tang2020onlineaugment}, \\    \\ LADA   \cite{kim2021lada}, \\ Ref. \cite{zhao2019data}, Ref. \cite{zhao2019data}\end{tabular} &
		\begin{tabular}[c]{@{}l@{}}Saves labor   (on constructing \\ transformation operations)\\    \\ Less   dependence on domain expertise\\    \\ Can model   subtle instances that may not\\  be observable to human developers\end{tabular} &
		\begin{tabular}[c]{@{}l@{}}Only   applicable in domains where\\  large data samples are accessible\\    \\ Time-consuming   training process\\    \\ May fail to   encode rare examples\end{tabular} \\
		\begin{tabular}[c]{@{}l@{}}Dynamic\\ augmentations\end{tabular} &
		\begin{tabular}[c]{@{}l@{}}AdaAug   \cite{cheung2021adaaug}, InstaAug \cite{miao2022learning}, \\ DeepAA   \cite{zheng2022deep}, \\ DivAug \cite{liu2021divaug}, SAPA   \cite{hu2021sapaugment}, \\ MetaAugment \cite{zhou2021metaA}\end{tabular} &
			\begin{tabular}[c]{@{}l@{}}More flexible and adaptive   augmentations\\    \\ Relatively wide range of   augmentations \\ possible   \\ More generic and easily transferable   \\ to different tasks and datasets\end{tabular} &
			\begin{tabular}[c]{@{}l@{}}Less   transparent\\    \\ Computationally   expensive\\    \\ Design process   not straightforward\end{tabular} \\
			\begin{tabular}[c]{@{}l@{}}Automatically generated\\  neural architectures (i.e. \\ NAS-based techniques)\end{tabular} &
			\begin{tabular}[c]{@{}l@{}}T-AutoML   \cite{yang2021t}, ARDA \cite{ chepurko2020arda},\\    \\ Ref.   \cite{lopes2021automl}, Tr-AutoML \cite{xue2019transferable},\\  Ref.   \cite{kashima2020joint}, MedPipe   \cite{chu2022medpipe}\end{tabular} &
			\begin{tabular}[c]{@{}l@{}}May complement   the traditional AutoML \\ data augmentation methods\\    \\ Can result in  more simple models with\\  powerful augmentation capabilities\\    \\ Can be used in   conjunction with other\\  approaches\end{tabular} &
			\begin{tabular}[c]{@{}l@{}}Generally   unintuitive\\    \\ Requires more   computational \\ resources\end{tabular} \\ \bottomrule
	
		\end{tabular}
            
        }
	\end{table*}

\subsection{NAS –based search space}

While most AutoML-based data augmentation methods typically learn effective data augmentation operations for fixed model architectures, an alternative approach to automated data augmentation  is to explore the search space for possible neural network architectures that perform well on the given data. These approaches (e.g., ARDA \cite{chepurko2020arda}, Ref. \cite{lopes2021automl}, T-AutoML \cite{yang2021t}, Tr-AutoML \cite{xue2019transferable}), instead of learning transformation functions to perform explicit data manipulation, are concerned with automatically generating the network itself. The goal of neural architecture search \cite{elsken2019neural} is to automate the process for creating ML models. In essence, it aims to select the best topology (i.e., network structure, including details of synaptic connections) for a given dataset and task. The approach is to take an input dataset and problem specification, specifically, in this case transformation of input images, and generate a model architecture that solves the given problem better than all other architectures for the given dataset. This involves finding hyperparameters that define the model structure (e.g., filter size and configuration, synaptic connection schemes, pooling and sampling details, model width and depth). The procedure is to iterate through model selection, where in each iteration the algorithm generates a different network structure, sets model parameters and validate its effectiveness based on performance on the target task. The process is completed when the best performing model is found. A large number of studies are devoted to this subject.
Some NAS-based automated data augmentation approaches have proposed to simultaneously find the best neural network architectures and effective augmentation policies to apply in order to achieve optimal results. The search space in this scenario consists of possible neural architectures, augmentation primitives and tunable hyperparameters. For instance, Kashima et al. in  \cite{kashima2020joint} proposed a NAS-based approach that aims to jointly find best NN architectures and effective augmentation policies that optimizes the performance of these architectures. Their approach combined FasterAA \cite{hataya2020faster} and DARTS \cite{liu2018darts} to jointly learn good neural architectures and the corresponding optimal data augmentation parameters. In this formulation, DARTS provides the set of possible model architectures while FasterAA implements the basic transformation operations that form the basis for generating data augmentation policies. Yang et al. \cite{yang2021t} argue that automating only some components of the ML task while hand-crafting others can result in sub-optimal performance. They introduce a transformer-based AutoML framework that aims to automate the entire deep learning pipeline – model architecture, data augmentation, as well as hyperparameter optimization. Thus, their approach jointly formulates (the problems of) neural architecture search (NAS), data augmentation, model training and evaluation. To achieve this, first, candidate architectures and augmentation operations, as well as the associated hyperparameters are represented as a one-dimentional vector.  A predictor is then trained to compare different combinations and configurations neural architectures and augmentations while finding the best hyperparameter values. 

One advantage of the NAS-based auto-augmentation approaches is that, compared to methods that perform explicit transformations (e.g., \cite{cubuk2019autoaugment,hataya2020faster,lin2021local, cubukpractical, lim2019fast}), they are able to generate far superior models that perform well on unseen datasets. However, the downside of most artificially generated models is that they tend to be more generic in nature and require relatively larger sizes, even for tasks where smaller models might suffice. Also, because of the need to train and validate all the possible neural architectures, the optimization process incurs very high computational costs and is unduly slow. Moreover, it requires huge amounts of storage resources since all candidate models need “live” for a little while in order to be trained and tested.

\section{Optimization of augmentation policies} \label{section:Optimization}

A naïve solution to the search problem is to test all possible augmentation operations and their possible combinations with different model configurations. Obviously, such a strategy is computationally costly and would be excessively time-consuming to practically realize.
Therefore, various optimization techniques are used to heuristically find the most effective combination of augmentations and their corresponding hyperparameters without carrying out exhaustive search (i.e., testing all possible augmentations). Many approaches utilize well-known optimization techniques – so-called black-box optimization methods – such as Bayesian optimization \cite{snoek2012practical}, evolutionary computation algorithms \cite{back1997handbook} and reinforcement learning \cite{kaelbling1996reinforcement}. These techniques are well suited for finding effective transformation operations and their parameters from a discrete space of primitive operations that cannot be described by analytical relationships. In this section, we present common search strategies, highlighting the main principles of operation, as well as the key strengths and weaknesses of each method.

The optimization process is an iterative procedure whereby the optimization algorithm is executed repeatedly while comparing the results of each iteration until an optimal or a satisfactory result is achieved. Although gradient-based methods have become the de facto approach for optimizing deep neural networks, the discontinuous nature of augmentation operations and their associated hyperparameters make it difficult to apply these techniques out-of-the-box. Heuristic search strategies are therefore the most viable approach for finding effective augmentation policies. classical optimization methods include random search, particle swarm optimization (PSO), ant colony optimization (ACO), simulated annealing, greedy algorithm, genetic algorithms and particle swarm optimization. The techniques have been developed for solving large-scale optimization problems in computer science. The popularity of these techniques stem from their ability to solve complex and ill-formalized AI problems for which analytical search or gradient-based methods are unsuited or do not provide the required level of performance. While a large diversity of heuristic search optimization techniques exist, only a small subset of the methods has been applied for the purpose of automating data augmentation. They include reinforcement learning, Bayesian Optimization (BO) and evolutionary computational algorithms. New search space design strategies have also been proposed that allow much simpler search techniques to produce good performance.  We discuss these approaches in this section.

\subsection{Reinforcement learning}

Reinforcement learning \cite{kaelbling1996reinforcement} is one of the first optimization techniques to be used in automated data augmentation \cite{cubuk2019autoaugment}.  It is a classical optimization technique for solving ill-formed problems and has wide applications in machine learning.  Indeed, a very large number of automated data augmentation approaches (e.g., AA \cite{cubuk2019autoaugment}, FastAA \cite{lim2019fast}, AWS \cite{tian2020improving}, PAA \cite{lin2021local} Learn2Augment \cite{gowda2022learn2augment}, Ref. \cite{chu2022augmentation} and RTS \cite{ruiz2018learning}) utilize RL for the optimization task. The reinforcement learning approach is based on the principle of an artificial agent interacting with the environment through specific permissible actions and making observations about the outcomes of different actions through rewards and penalties. In the context of data augmentation, the actions are specified as sequences of basic data transformation operations. Observations are the performance results based on a predefined evaluation criterion. Through repeated actions (i.e., application of the specified augmentation operations), the agent finds optimal augmentation strategies that maximize performance over time. In one of the pioneering works on AutoML-based data augmentation, Cubuk et al. \cite{cubuk2019autoaugment} utilize a reinforcement learning technique to simultaneously find good augmentations policies and optimal model hyperparameters that produced state-of-the-art performance on several benchmark datasets (see Figures \ref{fig:15CompCIFAR100} in Subsection \ref{Sec_AutoML-Performance}).  The authors \cite{cubuk2019autoaugment} utilized a recurrent neural network (RNN) as a controller in the RL formulation to dynamically predict the most effective data augmentation policy for a given dataset in the course of training. Their model, AutoAugment (AA) \cite{cubuk2019autoaugment}, searches for the best transformation operations, their probabilities, as well as the respective intensities with the help of the recurrent neural network (RNN)-based controller.

The approach is exceedingly expensive – even on such a relatively small dataset as CIFAR-10, it take about 5,000 GPU hours to train. Consequently, most subsequent works in this direction aim to achieve comparable performance while reducing the computational overheads. FasterAA \cite{hataya2020faster} and AWS  \cite{tian2020improving}, for example, propose to improve computational efficiency by employing weight-sharing mechanism within a reinforcement learning framework and achieve comparable results as AA on image classification tasks but with significantly lower computational budget. Patch AutoAugment (PAA) \cite{lin2021local} formulates the data augmentation optimization problem as cooperative multi-agent RL problem in which each aent learns an optimal augmentation policy for a small image patch. A multi-agent reinforcement learning algorithm based on decentralized partially observable Markov decision process \cite{white1993survey,spaan2012partially} is then used to learn a joint optimal augmentation policy over the entire image. 

Gowda et al. \cite{gowda2022learn2augment} propose a reinforcement learning approach to automatically learn to select the most useful video frames for augmentations on video datasets without actually executing and evaluating these augmentations.  The technique incorporates a so-called Semantic Matching sub-module to exploit the natural association of activities (i.e., foregrounds) and backgrounds. Augmentations are created by combining a pair of useful video segments, where one segment’s background is mixed with another’s foreground consisting of actor and objects. The approach achieves a performance gain of up to 4.4\% over classical augmentation methods.

Although many of the recent RL-based methods have significantly reduced computational cost of the optimization tasks, relative to other search methods, RL techniques are still more expensive for many practical purposes. Consequently, many alternative approaches are actively being explored.

\subsection{Bayesian Optimization}

\begin{figure}[H]
	\vspace {-1mm}
	\centering
	\includegraphics[width=1.0 \linewidth]{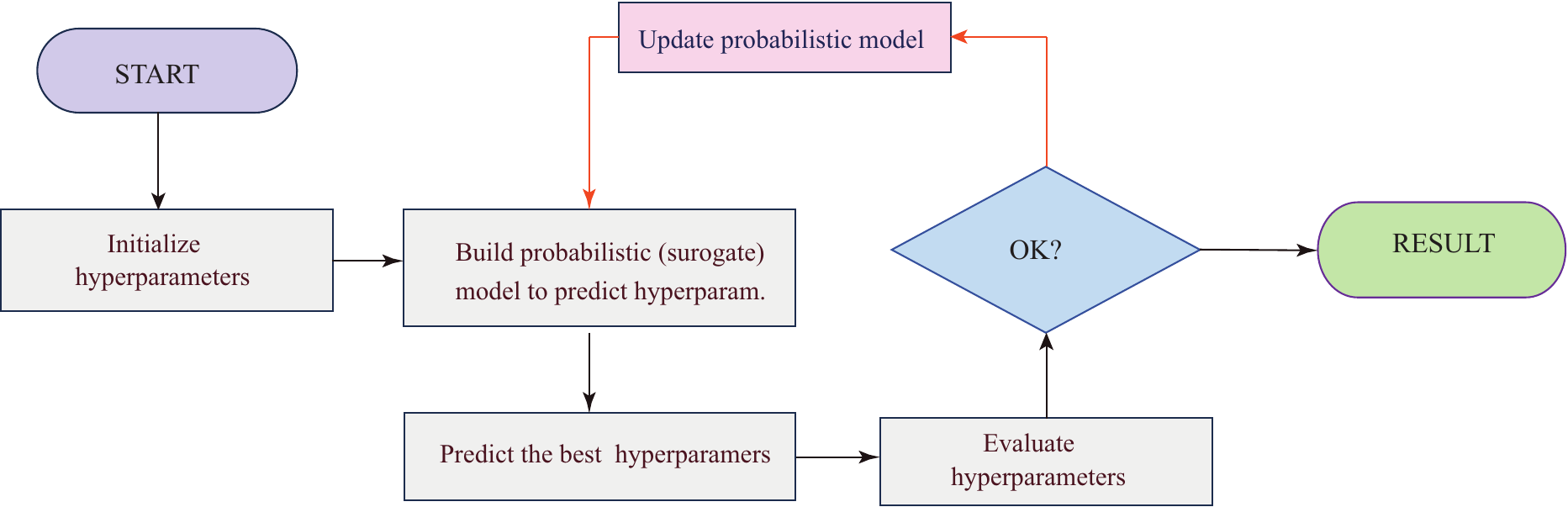} \vspace {-3mm}
	\caption{Simplified flowchart of Bayesian optimization. The approach relies on probabilistic modeling to predict the best augmentation hyperparametrs for a givn dataset and task. Based on these hyperparameters, a deep neural network is then trained and evaluated to establish the quality of the predicted hyperparameters.
	}\label{fig:11BO_AugPolicy}
\end{figure}

Because of the enormous computational requirements of RL methods such as AutoAugment and many follow-up works, recent approaches have focused on reducing computational requirements without incurring significant generalization performance penalty. To this end, a large number of techniques \cite{hataya2020faster,hu2021sapaugment,zhang2022bo} have been devised based on Bayesian optimization (BO) \cite{wang2023recent}. BO methods, as opposed to reinforcement learning approaches discussed in the previous subsection, allow to encode domain knowledge by keeping track of, and using past evaluation results to generate better policies in subsequent searches. For these approaches, the search space is formulated as a probability distribution in which subsequent searches are more focused on areas where the likelihood of the best hyperparameters lie. Thus, the BO approach implements a highly efficiency sampling strategy by selecting only the most promising set of hyperparameters to evaluate based on previous calls to the “evaluator”. A simplified workflow of the process is shown in Figure \ref{fig:11BO_AugPolicy}.

To further improve model efficiency, instead of training on the entire dataset, Fast AutoAugment (FastAA) \cite{lim2019fast} divides the training dataset into smaller subsets which are then trained simultaneously on different sub-models using BO strategy to find optimal augmentation policies. BO approaches have shown competitive results with relatively lower computational overheads compared to RL-based approaches due to the ability to estimate the quality of augmentations and skip bad augmentations policies before they are evaluated.

%%%%%%%%%%%%%%%% FIGURE 12: n %%%%%%%%%%%%%%%%%%%%%%
%\usepackage{float}
%%%%%%%%%%%%%%%%% End of Figure 11 %%%%%%%%%%%%%%%%%%%%

\subsection{Evolutionary computation algorithms}

\begin{figure}[H]
	\vspace {-1mm}
	\centering
	\includegraphics[width=1.0 \linewidth]{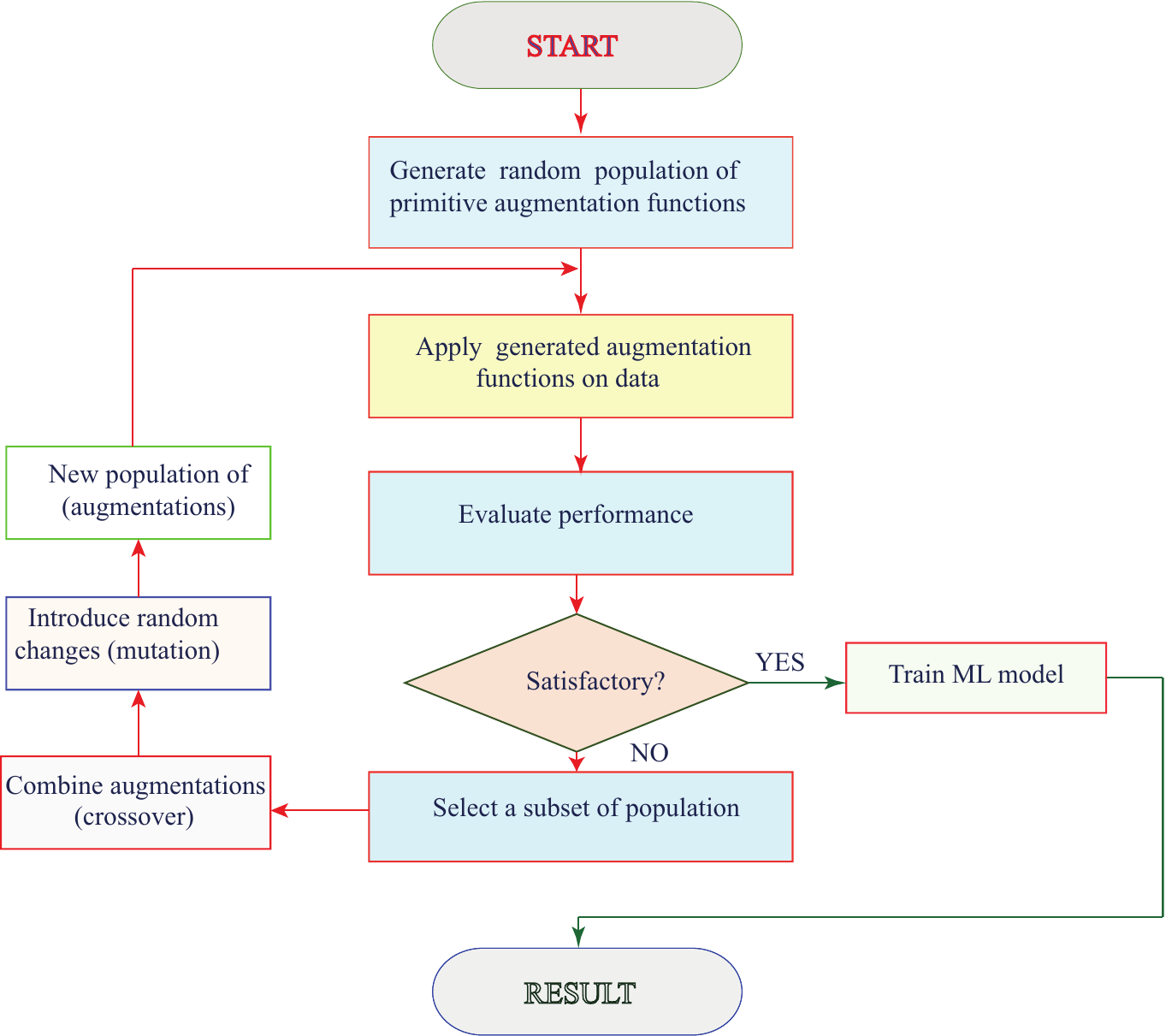} \vspace {-3mm}
	\caption{Basic principle of evolutionary computation algorithms. The is conecptually  similar to the procss of natural selection in living organisms.
	}\label{fig:12ECA_AugPolicy}
\end{figure}

As an alternative to the reinforcement learning and Bayesian optimization approaches, many works \cite{cheng2020improving} have proposed evolutionary computational techniques \cite{back1997handbook} for automating the search for effective data augmentation policies. Algorithms based on such techniques utilize specialized operators inspired by biological evolution processes—specifically, natural selection, crossover and mutation—for searching for, and optimizing the performance of useful augmentations in an iterative process during training (see Figure \ref{fig:12ECA_AugPolicy}). These techniques use the concept of natural selection to ensure that only high performing augmentation policies are maintained from one iteration to the next. The basic idea is to search from an initial pool of possible augmentations and select good transformation operations, and during each iteration of the search process to continually eliminate non-optimal transformations. The augmentations obtained may be combined with other augmentations in a process called crossover. Further, the population-based  technique periodically induces random changes in one or more augmentation operations of the current set by replacing some of the parameters with random parameters, thereby mutating the affected augmentations. Consequently, new and better set of augmentations are obtained by the process of crossover and mutation. Thus, in contrast to the other black-box optimization approaches discussed earlier that search over static transformation operations, approaches employing evolutionary computation algorithms allow the underlying augmentation operations to be varied and improved in the course of training. The enhancement is achieved by periodically making random changes to the current best policies, and inter-mixing to produce better characteristics. These operation are known in scientific literature as mutation and crossover, respectively. 
One of the first works to use evolutionary algorithm method for data augmentation was PBA \cite{ho2019population}. Subsequently, Cheng  et al. proposed a modified version of  PBA, Progressive Population Based Augmentation (PPBA) \cite{cheng2020improving}, by gradually reducing the search space in the course of training. This augmentation strategy was applied on multimodal data and showed good results.

%%%%%%%%%%%%%%%% FIGURE 13: n %%%%%%%%%%%%%%%%%%%%%%
%\usepackage{float}
%%%%%%%%%%%%%%%%% End of Figure 13 %%%%%%%%%%%%%%%%%%%%

When composing a search space for this approach, the main focus is to construct an initial set of useful transformations and a corresponding set of policies which can evolve with time. In this case, the augmentations can be composed using a variety of methods. For example, they can be sample space transformation operations (e.g., \cite{ho2019population}) or feature space (e.g., \cite{cheung2020modals}). Cheung et al. \cite{cheung2020modals} propose an evolutionary computation approach to find good augmentation policies composed in the latent space. Poor performing augmentations in the search space are gradually eliminated in the course of training while good ones are retained and enhance over time.

\subsection{Gradient-based methods}

Gradient-based automated data augmentation approaches are designed to approximate the gradients of augmentation hyperparameters rather than to iteratively search a discrete space for good augmentation policies. They are mainly aimed at achieving higher efficency compared with computational methods like RL and evolutionary algorithms

While gradient-based methods have proven to be universal and well adapted for training deep neural networks to find optimal parameters in modern machine learning algorithms, they are ill-suited for out-of-the-box AutoML-based data augmentation formulations. The application of gradient descent methods in solving automated data augmentation problems is constrained by the inherently non-differentiable nature of the search space. This is as a result of problems associated with discrete nature of transformation hyperparameters (e.g., discrete scale and rotation angle specifications), making it difficult to apply simple and effective gradient-based optimization techniques.

To enable the use of gradient descent methods for automated data augmentation, it is required that the search space be necessarily both continuous and monotonous, and that at each iteration in the search process, the direction of the greatest increase (or decrease) in the objective function can be determined until an optimal augmentation parameters are found. Several workarounds exist for ensuring that these conditions are met. Many of these approaches are based on previously developed techniques for solving general black-box optimization problems using the concept of approximate gradient estimation \cite{williams1992simple,grathwohl2017backpropagation,tucker2017rebar}. The approximate gradient estimator is essentially a differential operator that takes discrete parameter values or random, continuous variables and returns the corresponding gradients; it is a generalization of the concept of a gradient descent for non-differentiable and discrete functions. The use of this concept makes it possible to apply GD methods for finding optimal augmentation policies in a discrete search space.

Many approaches (e.g., DHA \cite{zhou2021dha}, DAAS \cite{wang2021daas}, AdvAA \cite{zhang2019adversarial}) invariably reformulate the search space in such a way that approximate numerical methods can be used for the computation of derivatives of the objective function. Faster AutoAugment \cite{hataya2020faster} uses straight-through estimator to simplify the representation of the search problem and allow the application of backpropagation to find optimal transformation operations. Similarly, instead of directly performing differentiation to estimate the gradient, MADAO \cite{hataya2022meta} data augmentation as a differentiable task using Neumann series approximation to compute implicit gradients for the objective function. Xu et al. \cite{xu2020automatic} employ stochastic relaxation method \cite{akimoto2019adaptive} to approximate the representation of non-differentiable augmentation hyperparameters in differentiable form to allow gradient-based policy optimization. 

Some new methods such as \cite{zhang2019adversarial,lin2019online}  are based on applying reinforcement learning concepts within a connectionist (i.e., network) framework to estimate the gradients of augmentation parameters. Zhang et al. \cite{zhang2019adversarial} proposed Adversarial AutoAugment (AdvAA) which computes approximate gradients with the aid of REINFORCE algorithm—a general statistical method based on reinforcement learning—proposed by Williams \cite{williams1992simple}. A similar approach is used in Online Hyper-parameter Learning for Auto-Augmentation (OHLAA) \cite{lin2019online} to estimate the gradient of the loss function with respect to the augmentation policy hyperparameters.

Different from the above methods, some approaches (e.g., \cite{li2020dada,shirobokov2020black}) rely on proxy models where augmentation variables can be represented as continuous and differentiable parameters, and used to indirectly optimize the discrete augmentation hyperparameters (through joint training). DADA \cite{li2020dada}, for example, proposes a continuous formulation of the discrete search space by introducing a continuous reparameterization of the discrete augmentation policy variables using a differentiable neural network as a surrogate model. The method is based on RELAX, one of several techniques proposed by Grathwohl et al. \cite{grathwohl2017backpropagation} to extend gradient descent to non-differentiable black-box optimization problems. It eliminates the need for continuous relaxation of the discrete variables by utilizing a surrogate neural network, whose parameters can be used to control the augmentation variables through joint optimization with the augmentation policy hyperparameters. Likewise, Shirobokov et al. \cite{shirobokov2020black} utilize differentiable surrogate neural networks that allow the optimization of simulator parameters for the synthesis of artificial data that are representative of real physical processes. We summarize the main strengths and weaknesses of the various optimization techniques covered in this work in Table \ref{tab:myOptimizationTechniques}.

\begin{table*}[h!]
			\caption{A comparison of the main strengths and weaknesses of the common optimization techniques used for automated data augmentation.}
		\label{tab:myOptimizationTechniques}
		
			\scalebox{0.95}{
				
		\begin{tabular}{@{}llll@{}}
			\toprule
			\textbf{Optimization method} &
			\textbf{Example works} &
			\textbf{Main strengths} &
			\textbf{Weaknesses} \\ \midrule
			Reinforcement learning &
			\cite{cubuk2019autoaugment,lim2019fast, lin2021local,gowda2022learn2augment} &
			\begin{tabular}[c]{@{}l@{}}Effective when cannot   be intuitively determined\\    \\ Does not require prior   knowledge about \\ augmentation operations\end{tabular} &
			\begin{tabular}[c]{@{}l@{}}Expensive to train\\    \\ Generally complex   architecture\end{tabular} \\
			Bayesian optimization &
			\cite{hataya2020faster,hu2021sapaugment,zhang2022bo} &
			\begin{tabular}[c]{@{}l@{}}More   flexible augmentation process\\    \\ Computationally   efficient\end{tabular} &
			\begin{tabular}[c]{@{}l@{}}Less transparent\\    \\ Computationally expensive\\    \\ Design process not straightforward\end{tabular} \\
			Evolutionary Algorithms &
			\cite{ho2019population,cheng2020improving,cheung2020modals} &
			\begin{tabular}[c]{@{}l@{}}Explicit mechanism to   adapt augmentation policies\\    \\ Evolution of policies   can lead to better than can be \\ intuitively designed\\    \\ No rigid upper ceiling   for performance\end{tabular} &
			\begin{tabular}[c]{@{}l@{}}Relatively expensive\\    \\ Relatively complex   schemes\end{tabular} \\
			Gradient methods &
			\cite{li2020dada,hataya2020faster,shirobokov2020black,lin2019online,hataya2022meta} &
			\begin{tabular}[c]{@{}l@{}}Fast training\\    \\ Simpler models\\    \\ Readily integrates with   DNN training process\end{tabular} &
			\begin{tabular}[c]{@{}l@{}}Often requires   approximations; can \\ compromise the fidelity of \\ representation\\    \\ Not always possible\end{tabular} \\
			Ensemble approaches &
			\cite{zhang2019adversarial,   hataya2020faster, luo2021autosmart,alaa2018autoprognosis} &
			\begin{tabular}[c]{@{}l@{}}Allows to exploit the   complementary benefits of \\ different methods\\    \\ Can generate more   diverse augmentations than \\ single methods\\    \\ Potentially more robust\end{tabular} &
			\begin{tabular}[c]{@{}l@{}}Can result in very   complex models\\    \\ Unintended interactions   may harm \\ performance\end{tabular} \\
			Searchless methods &
			\cite{cubukpractical,lingchen2020uniformaugment} &
			\begin{tabular}[c]{@{}l@{}}Fast training\\    \\ Relatively  simpler models\end{tabular} &
			\begin{tabular}[c]{@{}l@{}}Less flexible   augmentations\\    \\ Performance may have a   “hard” \\ upper ceiling\end{tabular} \\ \bottomrule
		\end{tabular}
		
     	}
	\end{table*}

\subsection{Ensemble optimization methods}

A recent hyperparameter search approach is to implement multiple optimization algorithms in a single AutoML model and then algorithmically determine the most effective ones for a given dataset and task. The ensemble search approach aims to leverage the strengths of different optimization methods to achieve higher performance by combining multiple search algorithms in a single framework. Since different optimization techniques are effective for different datasets and tasks \cite{feurer2018towards},  these approaches are particularly useful for AutoML models designed for generic applications that need to be able to handle multimodal data or process a broad range of data types for different machine learning problems.
 
Predictably, modern large-scale AutoML frameworks (e.g., TPOT \cite{olson2016tpot}, AutoDES \cite{zhao2022autodes}, Hyperopt-Sklearn \cite{komer2014hyperopt} and AutoWeka \cite{thornton2013auto}) typically rely on ensemble optimization techniques that can incorporate search strategies. In these frameworks, multiple optimization algorithms can be applied simultaneously. For instance, Auto-Tuned Models (ATM) \cite{swearingen2017atm} defines Bayesian optimization and reinforcement learning (multi-armed bandit) algorithms in its pipeline which can be used simultaneously to find optimal hyperparametters for the input data. Some models such as H2O AutoML\cite{ledell2020h2o} and AutoDES \cite{zhao2022autodes} incorporate a flexible representation of plausible pipeline structures that can be applied depending on the target task or input data type. Despite the increased model complexity and computational cost, the approach is often justified when solving broad machine learning problems involving very large and complex data.

\subsection{Searchless optimization of data augmentation policies}

Although the optimization techniques discussed in Subsections 5.1 through 5.4 have demonstrated impressive results, they are still too expensive for most practical purposes. For instance, optimization methods employed in AutoAugment takes about 15,000 GPU hours to  train on ImageNet \cite{zheng2022deep}. To address the challenge of computational complexity, some recent works (e.g., \cite{cubukpractical,naghizadeh2020greedy,muller2021trivialaugment,lingchen2020uniformaugment}) propose simplified search space design where good augmentation policies can be found using much simpler search strategies like random search \cite{ karnopp1963random}, grid search \cite{lavalle2004relationship} or greedy search \cite{wilt2010comparison}.  Greedy AutoAugment \cite{naghizadeh2020greedy} proposes to reduce the search space by eliminating the need to search for different combinations of augmentation operations in a combinatorial manner. The basic idea is to dynamically expand the space of augmentation operations in the direction of effective augmentation policies by creating and modifying sub-policies with good performance. In the search process, only transformation type and intensity are the only hyperparameteres considered.  The probability of applying the operations are determined later when the search is complete and is only applied on the effective policies found. This significantly reduces the amount of computations needed to find good augmentation strategies. Because of its simplicity and effectiveness, the greedy search policy optimization procedure has been used in several works \cite{momeny2022greedy}.

Approaches for further simplifying the search process have also been considered. In some cases, for example, UniformAugment (UA) \cite{lingchen2020uniformaugment}, these simplifications allow the search phase in the optimization process to be eliminated altogether without incurring significant performance penalty. RandomAugment (RA) \cite{cubukpractical} employs just two tunable parameters – augmentation intensity (i.e., transformation magnitudes) and selection probability—as variables to characterize the search space. Given the simplicity of this formulation, a simple grid search is sufficient to effectively search for optimal values of augmentation parameters. Instead of modeling selection probabilities as independent hyperparameters, in RA \cite{cubukpractical} the probability of applying any given augmentation operation is a constant value that is solely dependent on the number of applicable augmentation operations in the search space; it is computed as the inverse of the total number of operations. In the same way, the augmentation intensities are assumed to be uniform for all transformation operations. Based on this simplified reformulation of the search space, it is then possible to obtain optimal augmentation policies using only a single variable parameter, the augmentation intensity, and a constant—the total number of transformation operations which defines the application probability. RandomAugment RA \cite{cubukpractical}, is thus able to reduce the search space from 1032 possible operations to just 100. For this simple scenario, a naïve grid search provides very good results comparable to state-of-the-art performance. 

Unlike in RA \cite{cubukpractical} and all previous optimization strategy formulations, Uniform Augment (UA) \cite{lingchen2020uniformaugment} completely bypasses the search stage and instead propose to uniformly sample augmentation policy hyperparameters from the search space. To achieve this, the authors formulate the search space as distribution invariant by ensuring that the transformation operations and associated intensities result in label-preserving augmentations, and the transformed data remains within the distribution of the original input data. A machine learning model is then able to learn optimal augmentations by sampling randomly from the search space and optimizing model parameters using gradient descent. Uniform Augment, thus far, is the cheapest in terms of the number of computations needed to find optimal augmentation policies. Indeed, the number of search operations is theoretically zero.  

\section{Evaluation methods}
An important step after selecting a subset of operations for the search space and an optimization method is to evaluate the performance of the resulting models. This process is repeated for different combinations of model configuration settings (i.e., all plausible sets of augmentation operations, hyperparameters and search methods). At the end of the process, performance results are compared and an optimal pipeline and hyperparameters chosen according to a specified performance criterion. The costly nature of the evaluation stage significantly limits the  range of plausible configurations that can be  explored to achieve optimal predictive performance. Instead of training the network to converge before evaluating its performance, approaches have been devised to accelerate the process by enabling near-optimal strategies to be found without exhaustive evaluations.
 
 One of the most popular approaches is the so-called early stopping strategy \cite{he2021automl}, a technique based on terminating the evaluation process for settings that are predicted to perform relatively poor on a validation set. Some works, for instance, RandAugment (RA) \cite{cubukpractical}, TrivialAugment (TA) \cite{muller2021trivialaugment} and OHLAA \cite{lin2019online}, reduce the range of augmentation operations and hyperparameters in the search space in the evaluation process. A popular method to  speed up model evaluation is to reduce the fidelity of the input  data \cite{he2021automl}. In image augmentation, this is achieved by reducing the sizes or resolutions of the images used. Another common approach is to train on proxy tasks--i.e., employ reduced model size or subsets of training data. Examples of AutoML-based data augmentation models that employ this strategy include AutoAugment (AA) \cite{cubuk2019autoaugment}, FastAA \cite{lim2019fast}, \cite{lim2019fast}, FasterAA \cite{hataya2020faster}, DADA \cite{li2020dada}, AWS \cite{tian2020improving}, and PBA \cite{ho2019population}. Another method for reducing the enormous computaional overhead at the evaluation stage is to employ surrogate models as evaluators \cite{yao2018taking,zoller2021benchmark} to predict the performance of the target models. This approach circumvents the need to perform costly evaluation on different configurations of the real model.

\section{Quantitative performance of data augmentation approaches}

We present results on the performance of AutoML-based data augmentation methods and compare  these with state-of-the-art data augmentation methods based on classical approaches. For fairness we compare methods that use the same datasets and similar training settings (i.e., similar model configuration and number of training epochs). The results presented here are curated from the original works. We highlight instances where results or other performance information from secondary sources are used. 

 We first describe the datasets and settings used in most of the works. We also introduce the common performance benchmarks and evaluation metrics commonly used to compare the data augmentation methods investigated in this work. Finally, we present quantitative performance results and compare AutoML-based methods against several state-of-the-art tradditional methods. In addition to performance comparisons, we also show the effect of combining classical and AutoML-based augmentations. These results show that appreciable improvements can be gained by combining the two classes of methods.

\subsection{Datasets and model settings} \label{Sec_Label_datsets}

\subsubsection{Datasets} 
To assess the predictive performance of automated data augmentation methods, test results on the following datasets were considered: CIFAR-10 and CIFAR-100, SHVN \cite{netzer2011reading}, ImageNet \cite{deng2009imagenet}, and MS COCO \cite{lin2014microsoft}. We briefly each of these datasets in the next paragraphs.

The \textbf{CIFAR-10} \cite{krizhevsky2009learning} and \textbf{CIFAR-100} \cite{krizhevsky2009learning} are highly popular image classification datasets. Each of the datasets contains 60,000 labeled 32x32 RGB images that are divided into 50,000 training and 10,000 test images. The images in CIFAR-10 are divided into 10 classes of 6,000 images per class, while CIFAR-100 has 100 classes, each containing 600 (500 training and 10 test) images.

The \textbf{Street View House Numbers (SVHN)} \cite{netzer2011reading} is a digit (i.e., decimal numbers 0 to 9) recognition dataset containing 600,000 color (RGB) images. The dataset is made up of cropped, 32×32 images derived from Google Street View \cite{anguelov2010google}. Thus, the images are pictures of real-world house number plates in challenging settings.  SVHN include bounding box information in a separate file and can therefore be used for digit detection in addition to image classification.

\textbf{ImageNet} \cite{deng2009imagenet} is a large-scale visual recognition dataset for generic computer vision tasks. A subset of the dataset used by most works for image classification –ImageNet Large Scale Visual Recognition Challenge (ILSVRC) \cite{russakovsky2015imagenet}–has a total of 1,431,167 images divided into 1000 different object categories. The entire set is made up of 1,281,167 training, 100,000 test and 50,000 validation images. 

The \textbf{Microsoft Common Objects in Context (MS COCO)} dataset \cite{lin2014microsoft} is a large-scale dataset that can be used for many computer vision tasks. It is primarily designed for object detection, pose estimation, keypoint detection, image captioning, and segmentation tasks. It has a total of 330,000 images split into 80 classes.

\subsubsection{Performance metrics and model settings}

For the classification tasks, the performance metric commonly used is the generalization accuracy. Results for ImageNet include top-1\% and top-5\% accuracy measures. Mean average precision (mAP) is used to characterize performance of the object detection tasks, i.e., on the MS COCO dataset (Table \ref{tab:DetectionCOCO}).

Performance on CIFAR-10 and CIFAR-100 has been tested using three different backbone models: Wide Residual Networks \cite{zagoruyko2016wide}, specifically, Wide-ResNet-28-10 (see the original work for a detailed description of this setting); Shake-Shake (26 2x32d) \cite{gastaldi2017shake}; and PyramidNet \cite{lin2017feature}. Results for ImageNet dataset have are based on ResNet-50 and ResNet-200 backbone models (see He et al. \cite{he2016identity} for details). RetinaNet \cite{lin2017focal} is used with various  ResNet models \cite{he2016deep} on the MS COCO dataset.

\subsection{Performance of AutoML-based data augmentation methods} \label{Sec_AutoML-Performance}

\begin{figure}[H]
	\vspace {1mm}
	\centering
	\includegraphics[width=1.0 \linewidth]{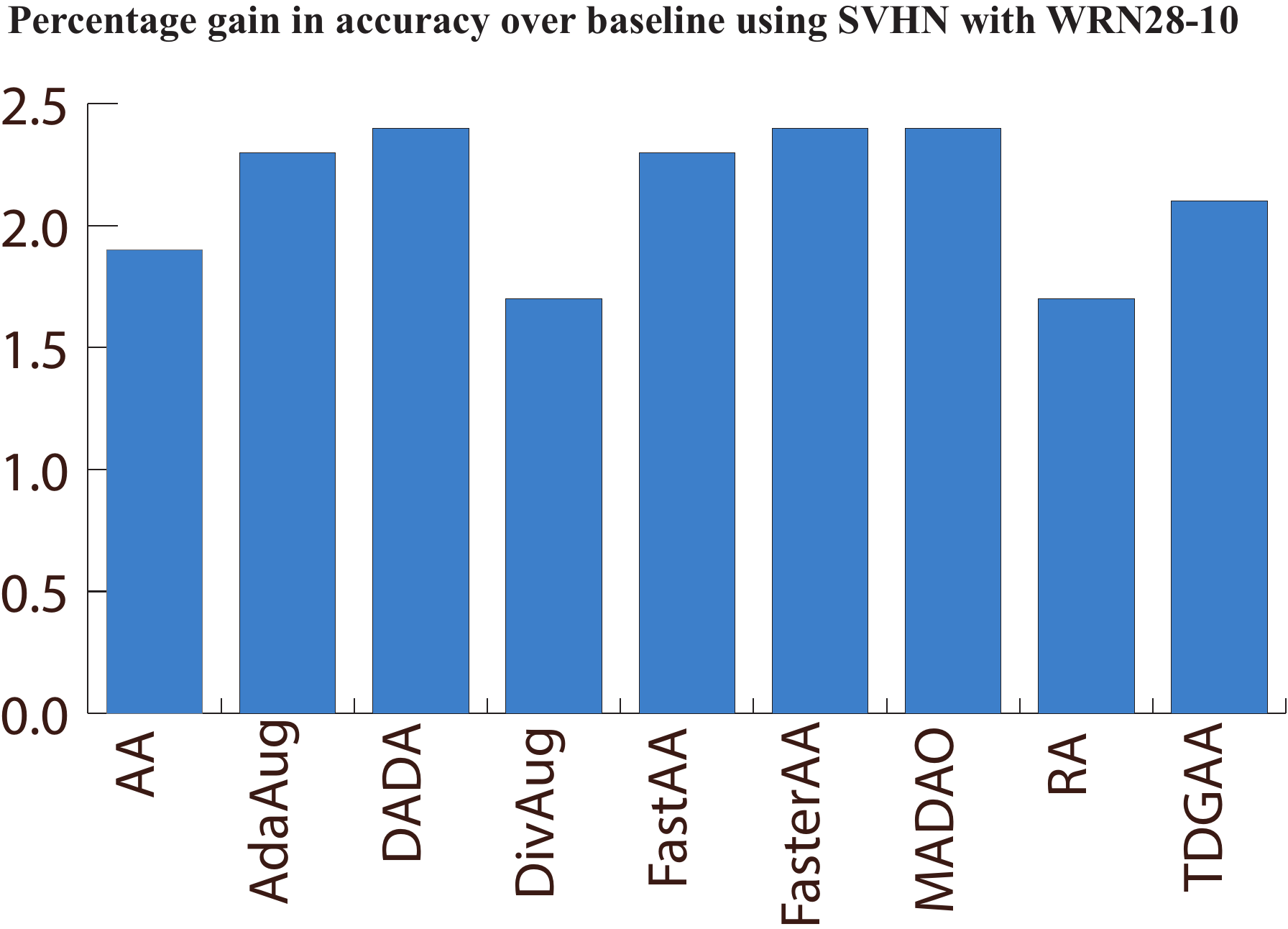} \vspace {-3mm}
	\caption{Digit recognition accuracy on the SVH dataset with Wide-ResNet-28-10 model. The results show performance gain, in percentage, of different AutoML-based data augmentation methods. 
	}\label{fig:13CompSVHN}
\end{figure}

Most of the existing works report results for image classification using large-scale CNN models. The most popular datasets used for these purposes are the ImageNet and CIFAR (CIFAR-10 and CIFAR-100) family of datasets described in Subsection \ref{Sec_Label_datsets}. Detailed performance results for automated data augmentation methods based on these datasets are summarized in Tables Tables \ref{tab:Tab_CIFAR}, \ref{tab:ImageNet} and \ref{tab:Avg-PerfTime}. Table \ref{tab:Tab_CIFAR} show image classification results for CIFAR-10 and CIFAR-10 for three different backbone models– WideResNet 28-10 (WRN28-10) \cite{zagoruyko2016wide}, Shake-Shake 26-32 (S-S 26-32) \cite{gastaldi2017shake} and PyramidNet (PNet) \cite{lin2017feature}. For each data augmentation method, we indicate the average classification accuracy for the three backbone models. From the Table, the mean gain in accuracy range between 0.6 and 1.5 on CIFAR-10 and between 0.8\% and 3.1\% on CIFAR-100. This demonstrates consistentently strong performance for all methods. Indeed, the average performance gain for all the methods is 1.02\% and 2.17\% on CIFAR-10 and CIFAR-100, respectively. Note that for CIFAR-100, many of the methods do not have results for PyramidNet backbone on CIFAR-100. We exclude these methods in our average performance computations. Similar to Table \ref{tab:Tab_CIFAR}, Table \ref{tab:ImageNet} shows classification results for the ImageNet dataset based on ResNet-50 and ResNet-200 backbone models. Results are shown for top-1 and top-5\% accuracies. The table also indicates percentage performance gain for each of these accuracy measures. The gains are consistent for all methods, with the lowest top-1\% accuracy gain  on RestNet-50 being +0.9\% (for PBA method) and +1.5\% on ResNet-200. The performance gains for various AutoML-based data augmentation methods on the SVHN dataset are summarized in Figure \ref{fig:13CompSVHN}. Again, the results show impressive performance on challenging digit classification tasks. In Table \ref{tab:Avg-PerfTime} we compute the average performance of different categories of automated data augmentation methods based on the optimization strategies they employ. The results show that performance difference across different search methods, on average, is largely insignificant.

While the majority of approaches have reported results for image classification tasks, a few works have evaluated the performance of automated data augmentation techniques on machine learning tasks other than image classification. For instance, Liu et al. \cite{liu2021direct}, Cubuk et al. \cite{cubukpractical} and Chen et al. \cite{chen2020gridmask} investigate the performance of common automated data augmentation methods–AA \cite{cubuk2019autoaugment}, DADA \cite{li2020differentiable}, DDAS \cite{liu2021direct}–on object detection tasks. The tests were conducted using large-scale detection models, specifically, ResNet-101 and RetinaNet. The results of these experiments are summarized in Table \ref{tab:DetectionCOCO}. These studies show that AutoML-based data augmentation methods are effective for object detection. Indeed, per the results in Table \ref{tab:DetectionCOCO}, the data augmentation techniques lead to an average performance improvement of 1.54\% (mean average precision or mAP) for object detection models.

\begin{table*}[h!]
	\caption{Performance of automated data augmentation methods on CIFAR-10 and CIFAR-100 datasets. The backbone models used are Wide-ResNet-28-10 (WRN 28-10), Shake-Shake-26 2x32d (S-S 26-32) and PyramidNet (PNet). The average recognition accuracy values over all three backbone models are indicated as mean.}
	\label{tab:Tab_CIFAR}

	\begin{tabular}{@{}lllllllll@{}}
		\toprule
		\multirow{2}{*}{\textbf{Method}}           & \multicolumn{4}{l}{\textbf{CIFAR-10}} & \multicolumn{4}{l}{\textbf{CIFAR-100}} \\ \cmidrule(l){2-9} 
		&
		\textbf{WRN28-10} &
		\textbf{\begin{tabular}[c]{@{}l@{}}S-S\\    \\ 26-32\end{tabular}} &
		\textbf{\begin{tabular}[c]{@{}l@{}}PNet\\    \\ +SD\end{tabular}} &
		\textbf{Mean} &
		\textbf{\begin{tabular}[c]{@{}l@{}}WRN\\ 28-10\end{tabular}} &
		\textbf{\begin{tabular}[c]{@{}l@{}}S-S\\    \\ 26-32\end{tabular}} &
		\textbf{PNet} &
		\textbf{Mean} \\ \cmidrule(r){1-1}
		Baseline                                   & 96.1     & 97.1    & 97.3   & 96.83   & 81.2    & 82.9    & 86.0    & 83.37    \\
		AutoAugment \cite{cubuk2019autoaugment}    & 97.4     & 98.1    & 98.5   & 98.0    & 82.9    & 85.7    & 89.3    & 85.97    \\
		AdaAug                                     & 97.4     &         &        &         &         &         &         &          \\
		AdvAA \cite{zhang2019adversarial}          & 98.1     & 98.2    & 98.6   & 98.3    & 84.5    & 85.9    & 89.6    & 86.67    \\
		AWS \cite{tian2020improving}               & 98.0     & 98.3    & 98.7   & 98.3    & 84.7    & 85.9    & 89.6    & 86.73    \\
		DADA \cite{li2020differentiable}           & 97.3     & 98.0    & 98.3   & 97.9    & 82.5    & 84.7    & 88.8    & 85.33    \\
		DDAS \cite{liu2021direct}                  & 97.3     & 97.9    & -      & 97.6    & 83.4    & 84.9    & -       & 84.15    \\
		DeepAA \cite{zheng2022deep}                & 97.4     & 98.1    & -      & 97.8    & 83.7    & 85.2    & -       & 84.45    \\
		DivAug \cite{liu2021divaug}                & 98.1     & 98.1    & 98.5   & 98.2    & 84.2    & 85.3    & -       & 84.75    \\
		FastAA \cite{lim2019fast}                  & 97.3     & 98.0    & 98.3   & 97.87   & 82.8    & 85.4    & 88.3    & 85.50    \\
		FasterAA \cite{hataya2020faster}           & 97.4     & 98.0    & -      & 97.70   & 82.2    & 84.4    & -       & 83.30    \\
		KeepAA                            & 97.8     & 97.8    & 97.8   & 97.8    & -       & -       & -       & -        \\
		MADAO                                      & 97.3     & -       & -      & 97.3    & -       & -       & -       & -        \\
		MetaAugment \cite{zhou2020metaaugment}     & 97.7     & 98.3    & 98.6   & 98.2    & 83.8    & 86.0    & 89.5    & 86.43    \\
		OLHA \cite{lin2019online}                  & 97.4     & -       & -      & 97.4    & -       & -       & -       & -        \\
		OnlineAugment \cite{tang2020onlineaugment}                       & 97.6     & -       & -      & 97.6    &         &         &         &          \\
		PAA \cite{lin2021local}                    & 97.8     & 98.2    & -      & 98.0    & 83.3    & -       & -       & 83.3     \\
		PBA \cite{ho2019population}                & 97.4     & 98.0    & 98.5   & 97.97   & 83.3    & 84.7    & 89.1    & 85.70    \\
		RA \cite{cubukpractical}                   & 97.3     & 98.0    & 98.5   & 98.0    & 83.3    & -       & -       & 83.30    \\
		RUA \cite{dong2021automating}              & 97.4     & -       & 98.5   & 97.4    & 83.6    & -       & -       & 83.6     \\
		TA \cite{muller2021trivialaugment}         & 97.5     & 98.2    & 98.6   & 98.1    & 84.3    & 86.2    & -       & 85.25    \\
		TeachAugment \cite{suzuki2022teachaugment} & 97.5     & 98.0    & 98.5   & 98.0    & -       & -       & -       & -        \\
		TDGA AA \cite{terauchi2021evolutionary}    & 97.14    & -       & -      & 97.14   & -       & -       & -       & -        \\
		UA \cite{lingchen2020uniformaugment}       & 97.3     & 98.1    & -   & 97.7    & 82.8    & 85.0    & -       & 83.90    \\
		OnlineAugment \cite{tang2020onlineaugment} & 97.6     & -       & -      & 97.6    & 83.4    & -       & -       & 83.40    \\ \bottomrule
	\end{tabular}
\end{table*}

\begin{table*}[h!]
	\caption{Performance of automated data augmentation methods on ImageNet dataset with ResNet-50 and ResNet-200 as backbone models. The gain in recognition accuracy is indicated as "Acc. improvement". }
	\label{tab:ImageNet}
	\begin{tabular}{@{}lllllllll@{}}
		\toprule
		\multirow{3}{*}{\textbf{Method}} &
		\multicolumn{4}{l}{\textbf{ResNet-50}} &
		\multicolumn{4}{l}{\textbf{ResNet-200}} \\ \cmidrule(l){2-9} 
		&
		\multirow{2}{*}{Top-1} &
		\multirow{2}{*}{Top-5} &
		\multicolumn{2}{l}{Acc. improvement} &
		\multirow{2}{*}{Top-1} &
		\multirow{2}{*}{Top-5} &
		\multicolumn{2}{l}{Acc. improvement} \\
		&      &      & Top-1 & Top-5 &      &      & Top-1 & Top-5 \\ \cmidrule(r){1-1}
		Baseline                                   & 76.3 & 93.1 & -     & -     & 78.5 & 94.2 & -     & -     \\
		AA \cite{cubuk2019autoaugment}             & 77.6 & 93.8 & +1.3  & +0.7  & 80.0 & 95.0 & +1.5  & +0.8  \\
		AdvAA \cite{zhang2019adversarial}          & 79.9 & 94.5 & +3.6  & +1.4  & 81.3 & 95.3 & +2.8  & +1.1  \\
		AWS \cite{tian2020improving}               & 79.4 & 94.5 & +3.1  & +1.4  & 81.4 & 95.3 & +2.9  & +1.1  \\
		DADA \cite{li2020differentiable}           & 77.5 & 93.5 & +1.2  & +0.4  & -    & -    & -     & -     \\
		FastAA \cite{lim2019fast}                  & 77.6 & 93.7 & +1.3  & +0.6  & 80.6 & 95.3 & +2.1  & +1.1  \\
		OLHA \cite{lin2019online}                  & 78.9 & 94.3 & +2.6  & +1.1  & -    & -    & -     & -     \\
		OnlineAugment [OAug]                       & 77.6 & -    &       &       &      &      &       &       \\
		MetaAugment   \cite{zhou2020metaaugment}   & 79.7 & 94.6 & +3.4  & +1.5  & 81.4 & 95.5 & +2.9  & +1.3  \\
		PAA \cite{lin2021local}                    & 77.5 & -    & +1.2  & -     & -    & -    & -     & -     \\
		PBA \cite{ho2019population}                & 77.2 & 93.4 & +0.9  & +0.3  & -    & -    & -     & -     \\
		TeachAugment                               & 77.8 & 93.7 &       &       &      &      &       &       \\
		RA \cite{cubukpractical}                   & 77.6 & 93.8 & +1.3  & +0.7  & -    & -    & -     & -     \\
		RUA \cite{dong2021automating}              & 77.7 & -    & +1.4  & -     & -    & -    & -     & -     \\
		OnlineAugment \cite{tang2020onlineaugment} & 77.5 & -    & +1.2  & -     & -    & -    & -     & -     \\
		TA \cite{muller2021trivialaugment}         & 78.1 & 93.9 & +1.8  & +0.8  & -    & -    & -     & -     \\ \bottomrule
	\end{tabular}
\end{table*}

\begin{table*}[h!]
	\caption{Object detection performance (in mean average precision) of different automated augmentation methods on MS COCO dataset. Results have been compiled from performance reports of various works. These are: \cite{liu2021direct,chen2020gridmask,cubukpractical}.}
	\label{tab:DetectionCOCO}
	\begin{tabular}{@{}lllll@{}}
		\toprule
		\textbf{Backbone model} &
		\textbf{Aug. Method} &
		\textbf{mAP} &
		\textbf{Gain in mAP} &
		\textbf{Results from} \\ \midrule
		\multirow{4}{*}{\begin{tabular}[c]{@{}l@{}}ResNet-50 + RetinaNet\end{tabular}} &
		Baseline &
		36.9 &
		0 &
		\multirow{4}{*}{Ref. \cite{liu2021direct}} \\
		&
		DADA &
		38.4 &
		1.5 &
		\\
		&
		DDAS &
		38.1 &
		1.2 &
		\\
		&
		AA &
		- &
		- &
		\\
		\multirow{4}{*}{\begin{tabular}[c]{@{}l@{}}ResNet-101 + RetinaNet\end{tabular}} & Baseline & 38.6 & 0 & \multirow{4}{*}{Ref. \cite{chen2020gridmask}} \\
		&
		DADA &
		40.0 &
		1.4 &
		\\
		&
		DDAS &
		40.1 &
		1.5 &
		\\
		&
		AA &
		39.8 &
		1.2 &
		\\
		\multirow{3}{*}{\begin{tabular}[c]{@{}l@{}}ResNet-101 + RetinaNet\end{tabular}} &
		Baseline &
		38.8 &
		0 &
		\multirow{3}{*}{Ref. \cite{cubukpractical}} \\
		&
		AA &
		40.4 &
		1.6 &
		\\
		&
		RA &
		40.1 &
		1.3 &
		\\
		\multirow{3}{*}{\begin{tabular}[c]{@{}l@{}}ResNet-200 + RetinaNet\end{tabular}} &
		Baseline &
		39.9 &
		0 &
		\multirow{3}{*}{Ref. \cite{cubukpractical}} \\
		&
		AA &
		42.1 &
		2.2 &
		\\
		&
		RA &
		41.9 &
		2.0 &
		\\ \cmidrule(lr){1-5}
	\end{tabular}
\end{table*}

%%%%%%%%%%%%%%%% FIGURE 13   %%%%%%%%%%%%%%%%%%%%%%
%\usepackage{float}
%%%%%%%%%%%%%%%%% End of Figure 13 %%%%%%%%%%%%%%%%%%%%

%%%%%%%%%%%%%%%% FIGURE 14 %%%%%%%%%%%%%%%%%%%%%
%\usepackage{float}
\begin{figure*}[!htb]
	\vspace {3mm}
	\centering
	\includegraphics[width=0.80 \linewidth]{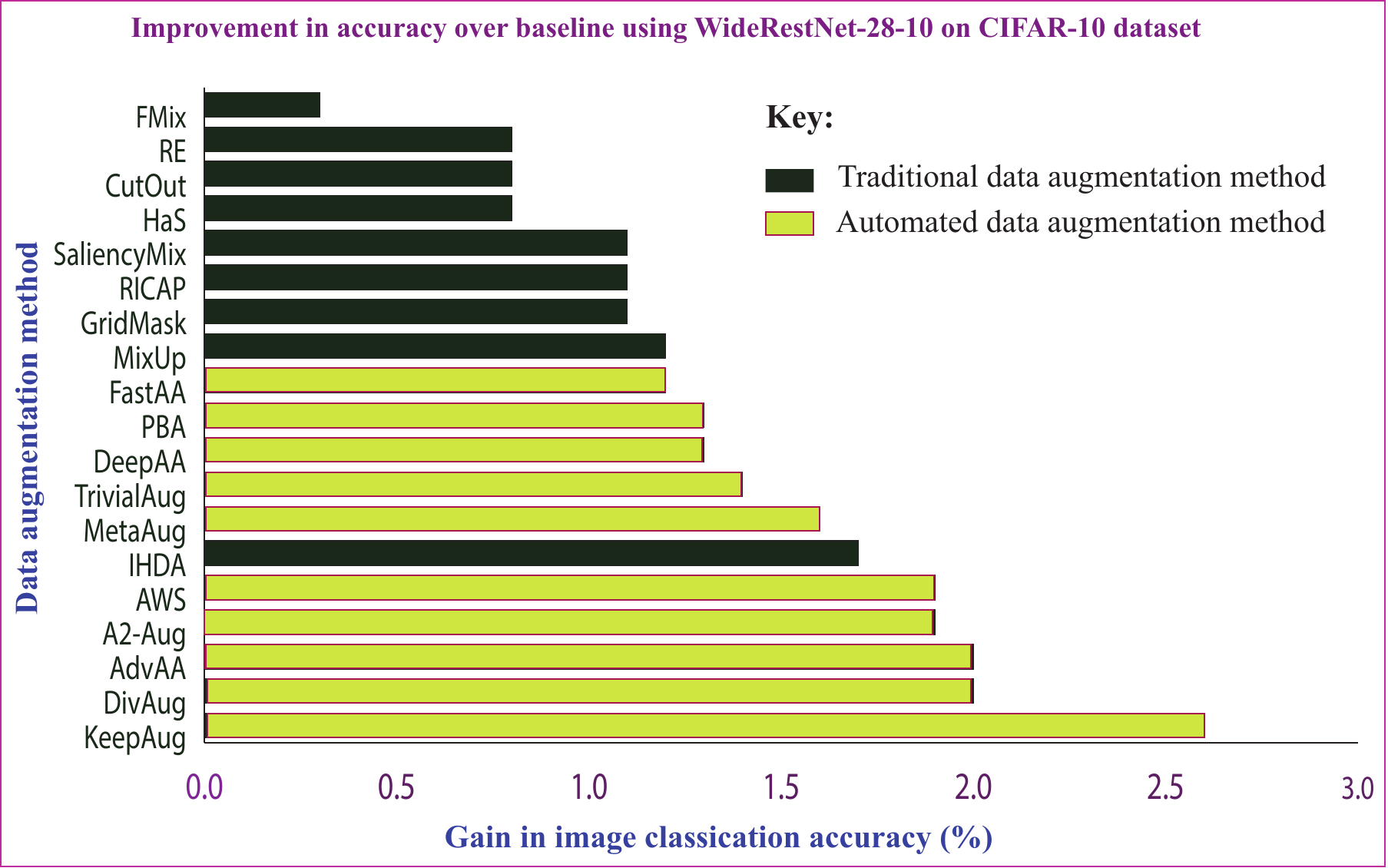} \vspace {-3mm}
	\caption{A comparison of gain in predictive performance (percentage accuracy over baseline) of automated data augmentation methods and augmentation techniques based on conventional paradigms. The baseline accuracy is 96.1\%. 
	}\label{fig:14CompCIFAR10}
\end{figure*}
%%%%%%%%%%%%%%%%% End of Figure 11 %%%%%%%%%%%%%%%%%%%%

%%%%%%%%%%%%%%%% FIGURE 15 %%%%%%%%%%%%%%%%%%%%%
%\usepackage{float}
\begin{figure*}[!htb]
	\vspace {-3mm}
	\centering
	\includegraphics[width=0.80 \linewidth]{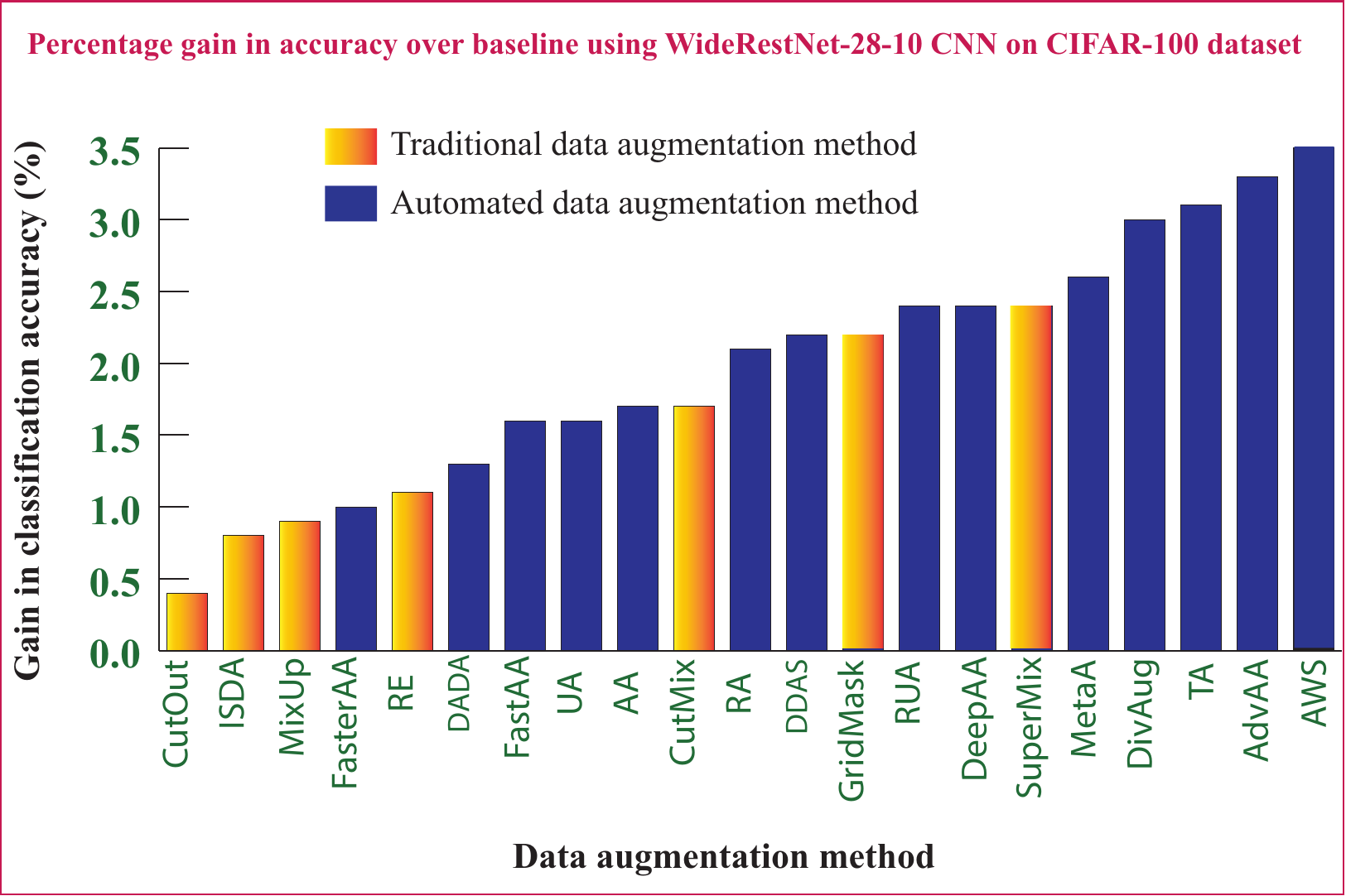} \vspace {-3mm}
	\caption{Performance of automated data augmentation techniques versus manual methods on CIFAR-100 using Wide-ResNet-28-10 CNN backbone. The results depict improvement gain (in \% accuracy) over the baseline. Baseline performance is81.2\%.
	}\label{fig:15CompCIFAR100}
\end{figure*}
%%%%%%%%%%%%%%%%% End of Figure 15 %%%%%%%%%%%%%%%%%%%%

%%%%%%%%%%%%%%%% FIGURE 16   %%%%%%%%%%%%%%%%%%%%%%
%\usepackage{float}
 \begin{figure*}[h]
	\vspace {3mm}
	\centering
	\includegraphics[width=0.75 \linewidth]{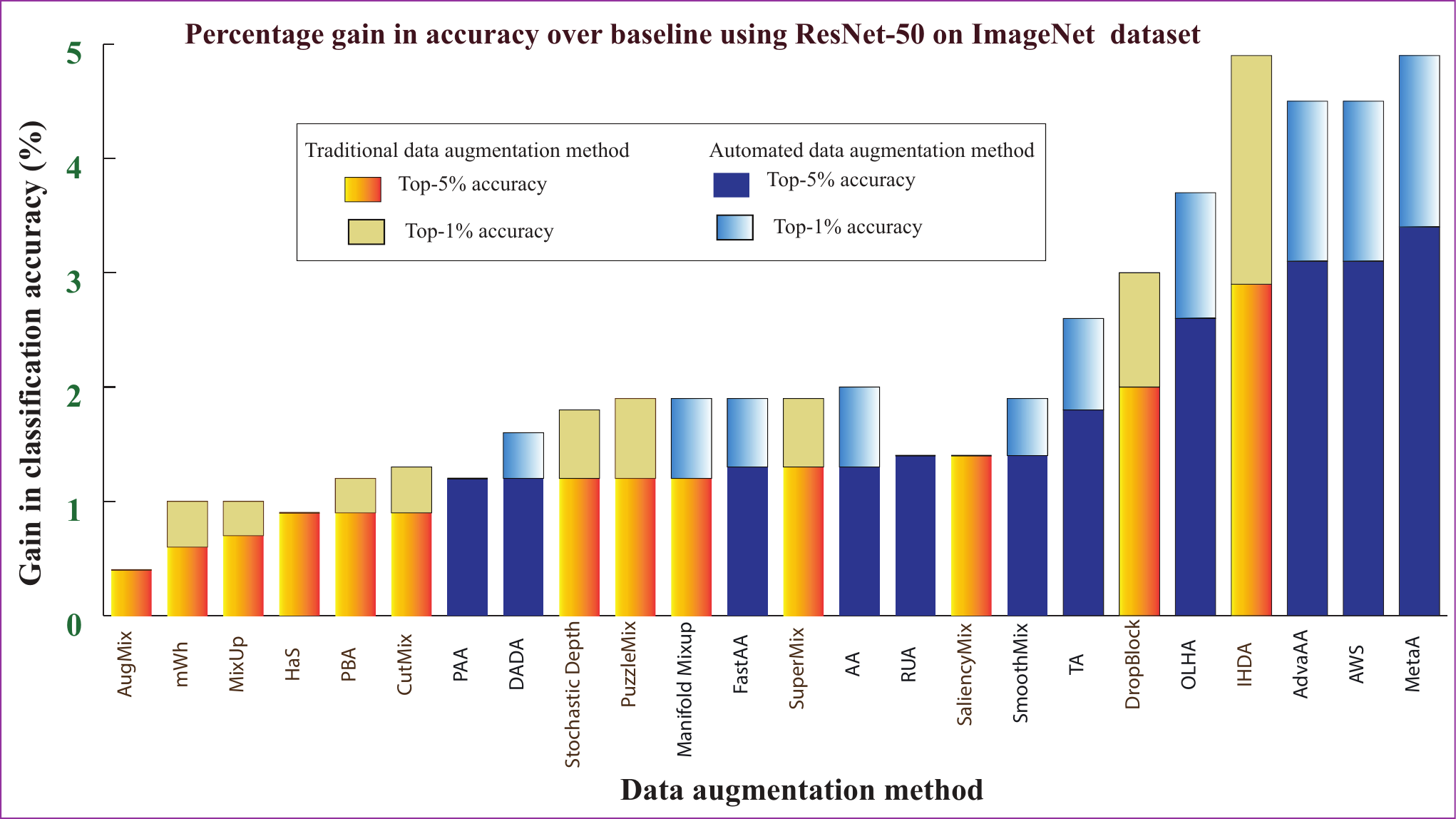} \vspace {-3mm}
	\caption{Comparison of gain in performance for automated data augmentation and conventional methods on ImageNet dataset using ResNet-50. The baseline performance is 76.3\% and 93.1\% for top-1 and top-5\% accuracy, respectively.
	}\label{fig:16CompareImageNet}
\end{figure*}
%%%%%%%%%%%%%%%%% End of Figure 16 %%%%%%%%%%%%%%%%%%%%

\subsection{Comparision of performance of classical and AutoML-based data augmentation methods}

\subsubsection{Classical data augmentation methods}

The performance of many state-of-the-art classical data augmentation methods have also been evaluated for the considered benchmark datasets (i.e., CIFAR-10, CIFAR-100 and ImageNet) and model settings. The traditional augmentation methods compared on ImageNet include CutMix \cite{yun2019cutmix}, SuperMix \cite{dabouei2021supermix}, MixUp \cite{zhang2017mixup}, StochasticDepth \cite{huang2016deep}, ISDA \cite{wang2019implicit}, Manifold Mixup \cite{verma2019manifold}, PuzzleMix \cite{kim2020puzzle}, DropBlock, IHDA \cite{khan2020post} and SaliencyMix \cite{uddin2020saliencymix}. For CIFAR-10 and CIFAR-100, the following methods are covered: MixUp \cite{zhang2017mixup}, CutOut \cite{devries2017improved}, CutMix \cite{yun2019cutmix}, ISDA \cite{wang2019implicit}, AgMax \cite{atienza2022improving} and Random Erase (RE) \cite{zhong2020random}. The performance of these works are reported in their original sources and in various surveys. Using those results, we compute the performance gains for various datasets and models, and compare with the performance of automated data augmentation methods. These results are discussed in Subsection \ref{Sec_comparison_AutoDA_vsTradDA}.

\subsubsection{Comparison of automated data augmentation methods and state-of-the art classical approaches} \label{Sec_comparison_AutoDA_vsTradDA}

In this section, we compare the performance of state-of-the-art data agumentation methods based on classical approaches with automated data augmentation methods. The results are summarized in Figures \ref{fig:14CompCIFAR10}, \ref{fig:15CompCIFAR100} and \ref{fig:16CompareImageNet}. Again, to ensure a uniform point of reference for the assessment, we selected methods that have been evaluated on the same datasets and CNN backbone models for comparison. Specifically, the results provided for CIFAR-10 and CIFAR-100 have been tested with Wide-ResNet-28-10, Shake-Shake 26 2x32d and PyramidNet backbones; SVHN with Wide-ResNet-28-10 backbone; and ImageNet with ResNet-50 and ResNet-200 backbones. 

For the sake of clarity, we computed and compared the gain in accuracy for all the methods based on quantitative results reported in the various original works. Overall, the results show consistently high performance for automated data augmentation methods over classical methods. In Figure \ref{fig:14CompCIFAR10}, we compare the performance of the best automated data augmentation methods with state-of-the-art approaches based on classical techniques on CIFAR-10. In the case of classical methods, IHDA \cite{khan2020post} shows the highest performance (+1.7\%), followed by MixUP (+1.2\%) and GridMask (+1.1\%). These are all well-below the best automated augmentation methods, which achieve an average of +2.17\% gain in classification accuracy. In Figure \ref{fig:14CompCIFAR10}, it can be observed that only one classical data augmentation method, IHDA \cite{khan2020post} at \[{6^{th}}\], ranks among the top ten data augmentation methods. Similarly, for CIFAR-100 (Figure \ref{fig:15CompCIFAR100}), the automated methods have convincingly outperformed the classical augmentation methods, averaging +1.47\% gain in performance compared with +2.28\% for classical methods. Only two classical methods–SuperMix(+2.4\%) and GridMask (+2.2\%)–are part of the ten best performing augmentation methods. Figure \ref{fig:16CompareImageNet} shows the relative performance of automated and trational methods on ImageNet dataset with ResNet-50 backbone. Results are shown for top-1\% and top-5\% accuracies. Here, too, AutoMl-based augmentation methods significantly outperform their classical counterparts. For instance, for the top-5\% classification accuracies, four out of the top five and eight out of the top ten methods are automated augmentation methods. These results convincingly demonstrate the superior performance of data augmentation techniques based AutoML pipelines.

\subsubsection{Improving performance by combining classical and AutoML methods}

Combining automated data augmentation strategies with multiple classical data augmentation methods has been shown by several authors (e.g., Atienza \cite{atienza2022improving} and Tang et al. \cite{tang2020onlineaugment}) to be effective in image classification tasks. Atienza \cite{atienza2022improving} compared the gain in classification accuracy from AutoAugment (AA) and several other state-of-the-art classical data augmentation techniques (specifically, CutMix \cite{walawalkar2020attentive}, CutOut \cite{devries2017improved}, MixUp \cite{zhang2017mixup} and AgMax \cite{atienza2022improving}) with a deep-learned augmentation strategy (AA \cite{cubuk2019autoaugment}).

They first assessed the performance of each technique separately before pairing AutoAugment with each of the classical augmentation methods. The results obtained by their experiments showed marked improvement by the paired augmentation strategies over AutoAugment. The combined augmentations also outperformed each of the classical methods when applied independently. All cases of combined augmentations resulted in significant performance improvements over corresponding single augmentations. More importantly, utilizing a combination of three augmentations also showed a marked improvement over those employing only two augmentation strategies. The results of the study are summarized in Table \ref{tab:CombinedAA_Trad}. The author used CIFAR10 dataset with WideResNet-40-2 backbone in their experimental setting. Note that the original works AA, MixUp and CutMix do not report results for the WideResNet-28-10 backbone. Because of the difference in training settings, the accuracies of individual methods reported (and shown in the Figure \ref{fig:14CompCIFAR10}) are slightly different from those obtained by the original sources. Nonetheless, they reflect the general performance of the models based on the specific techniques and dataset configurations. The experimental results show that while automated data augmentation methods hold significant promise, there is still scope for the application of classical methods to complement augmentations generated automatically. However, applying multiple augmentations in this manner ought to be carefully considered as combining augmentations have been shown to harm performance in some settings (e.g., \cite{wen2020combining}).

% \usepackage{booktabs}
\begin{table}[]
	\caption{Performance gain by combining classical and AutoML-based data augmentation methods. The models were tested according to the specific settings described by Atienza \cite{atienza2022improving}.}
	\label{tab:CombinedAA_Trad}
	\begin{tabular}{@{}lll@{}}
		\toprule
		\textbf{\begin{tabular}[c]{@{}l@{}}Data Augmentation   \\    \\ method\end{tabular}} & \textbf{mAP} & \textbf{Acc.   Gain} \\ \midrule
		\multicolumn{3}{l}{AA and   classical methods applied separately}  \\
		Standard   (no aug)                      & 95.1           & o        \\
		AA \cite{cubuk2019autoaugment}           & 95.9           & 0.8      \\
		CutMix \cite{yun2019cutmix}              & 96.2           & 1.1      \\
		CutOut \cite{devries2017improved}        & 96.2           & 1.1      \\
		MixUp \cite{zhang2017mixup}                 & 95.8           & 0.7      \\
			AgMax \cite{atienza2022improving}        & 95.6           & 0.5      \\
			\multicolumn{3}{l}{Pairing of AA   with one classical method}      \\
			AA + CutOut                              & 96.4           & 1.3      \\
			AA + MixUp                               & 96.0           & 0.9      \\
			AA + CutMix                              & 96.4           & 1.3      \\
			AA[5] + AgMax                            & 96.4(0.5)      & 1.3      \\
			\multicolumn{3}{l}{Pairing of  classical methods}                  \\
			CutMix + AgMax                           & 96.7(0.5)      & 1.6      \\
			CutOut  + AgMax                          & 96.6(0.4)      & 1.5      \\
			MixUp + AgMax                            & 96.3(0.5)      & 1.2      \\
			\multicolumn{3}{l}{combination   of  two classical methods and AA} \\
			CutOut+AA + AgMax                        & 97.1(0.7)      & 2.0      \\
			MixUp+AA + AgMax                         & 96.6(0.6)      & .5       \\
			CutMix+AA + AgMax                        & 96.8(0.4)      & 1.7      \\ \bottomrule
		\end{tabular}
	\end{table}

	%%%%%%%%%%%%%%%%%%%%%%%%%%%%%%%%%%%%%%%%%%%%%%%%%%%%%%%%%%%%%%%%%%%%%%%%
	%%%%%%%%%%%%%%%%%%%%%%%%%%%%%%%%%%%%%%%%%%%%%%%%%%%%%%%%%%%%%%%%%%%%%%%%

	\begin{table*}[h!]
		\caption{Average performance of different search strategies for automated data augmentation}
		\label{tab:Avg-PerfTime}
		\begin{tabular}{@{}llll@{}}
			\toprule
			\textbf{Search strategy}                                                 & \textbf{Example works}                                                                                                                                                                            & \begin{tabular}[c]{@{}l@{}} \textbf{Acc.*}\\    \textbf{(Avg)}\end{tabular} & \begin{tabular}[c]{@{}l@{}} \textbf{Time**} \\ \textbf{(Avg)}\end{tabular} \\ \midrule
			\begin{tabular}[c]{@{}l@{}}Reinforcement learning (RL)\end{tabular}   & \begin{tabular}[c]{@{}l@{}}AA\cite{cubuk2019autoaugment}, AWS \cite{tian2020improving}, PAA \cite{lin2021local}, Ref. \cite{chu2022augmentation}, FasterAA \cite{hataya2020faster}, BDA \cite{lu2023bda}\end{tabular} & 78.2                                                                & 210   \\                                                           \\
			\begin{tabular}[c]{@{}l@{}}Evolutionary algorithm (EA)\end{tabular}   & \begin{tabular}[c]{@{}l@{}}PBA \cite{ho2019population}), MODALS \cite{cheung2020modals}, PPBA \cite{cheng2020improving}, TDGA \cite{terauchi2021evolutionary}\end{tabular}                     & 77.2                                                                & 42     \\                                                          \\
			\begin{tabular}[c]{@{}l@{}}Gradient descent (GD)\end{tabular}         & \begin{tabular}[c]{@{}l@{}}DADA \cite{li2020dada}, AutoDO \cite{gudovskiy2021autodo}, OHLAA \cite{lin2019online}, AdvAA \cite{zhang2019adversarial}\end{tabular}                               & 78.8                                                                & 3.5                  \\                                            \\
			\begin{tabular}[c]{@{}l@{}}Bayesian optimization \end{tabular} & \begin{tabular}[c]{@{}l@{}}SAPA \cite{hu2021sapaugment}, BO-Aug \cite{zhang2022bo}, FastAA \cite{lim2019fast}\end{tabular}                  & 77.6                                                                & 6.3           \\                                                   \\
			\begin{tabular}[c]{@{}l@{}}Greedy search and random search\end{tabular}  & GreedyAA \cite{naghizadeh2020greedy}   Ref. \cite{momeny2022greedy,naghizadeh2020greedy}                                                                                                          & 77.5                                                                & 8.8          \\                                                    \\
			Searchless methods                                                       & RA \cite{cubukpractical}, UA \cite{lingchen2020uniformaugment}                                                                                                                                    & 77.6                                                                & 0.0***                                                           \\ \bottomrule
		\end{tabular}
	\end{table*}
	%%%%%%%%%%%%%%%%%%%%%%%%%%%%%%%%%%%%%%%%%%%%%%%%%%%%%%%%%%%%%%%%%%%%%%%5
	%%%%%%%%%%%%%%%%%%%%%%%%%%%%%%%%%%%%%%%%%%%%%%%%%%%%%%%%%%%%%%%%%%%%%%5
	
\section{Discussions}

\subsection{The importance of computational efficiency}

Because of the enormous compute resource demand of earlier approaches, most works (e.g., FasterAA, PAA AdvAA, OHLAA and AWS) have focused more on balancing generalization accuracy and computational efficiency. These works have achieved comparable performance to AutoAugment, the original approach proposed by Cubuk et al. \cite{cubuk2019autoaugment} while reducing the computational overhead. Many studies investigate more efficient optimization techniques in order to overcome the inherent computational complexity of RL methods. For instance, results of DDAS, FastAA and RA all demonstrate equivalent or slightly lower generalization accuracy than AA \cite{cubuk2019autoaugment}—one of the original works that introduced the concept of data augmentation based AutoML—but significantly reduced the computation time from 5,000 GPU hours in AA to just 10 (e.g., 5 hours in PBA, 3.5 in FastAA, and 0.23 hours in FasterAA) .
Some other works [e.g., UA and RA) focus on simplifying the search space so that good augmentation policies can be found using much simpler search strategies or without explicitly conducting search. These approaches, despite the significant reduction in compute time, have shown competitive performance. However, there seems to be little room to further extend their predictive performance.

\subsection{Predictive performance versus computational requirements }

As stated earlier, a major problem with AutoML-based data augmentation approaches is the tendency to excessively expand the training set in a bid to increase predictive performance. This can lead to situations where small performance gains are achieved at the expense of a disproportionately high increase in model complexity and computational resource requirements. Consequently, in practical settings, there is often a trade-off between model predictive accuracy on one hand, and computational and space complexity and interpretability on the other. Some state-of-the-art methods such as \cite{tsamardinos2022just} and \cite{xanthopoulos2020putting} provide mechanisms to set priority levels to achieve the right balance of predictive performance, computational budget and model interpretability. For instance, Tsamardinos et al. \cite{tsamardinos2022just} design different model configuration options, with each configuration prioritizing a specific objective: interpretability, predictive performance, and minimization of model size through more aggressive feature reduction. These configuration settings allow users to customize the AutoML pipeline according to their particular needs and constraints.  This capability is widely used in popular AutoML tools that are designed for generic applications. 

\subsection{Open problems and research directions}

Automated data augmentation methods have in recent years shown a lot of promise and, as confirmed by the comparative quantitative results in Section 7, have outperformed state-of-the-art manual data augmentation techniques on various benchmarks. Despite the impressive feat, there are a number of important challenges and unsolved problems. We summarize some of the key ones here.

\begin{itemize}

	\item One of the most challenging tasks in the implementation of automated data augmentation, and indeed in the development of AutoML models in general, is the task of composing a good search space that can provide effective solution for the target application. Even though approaches that learn to automatically generate effective search spaces from data have been devised, they generally require substantial domain knowledge to design, making it challenging for nonexperts to accomplish. Recently, large language models (LMMs) have been leveraged to automate complex machine learning tasks. Preliminary work by Yang et al. \cite{yang2024autommlab} has shown that with LLMs it will be possible to fully automate the development of AutoML and operation of frameworks in a way that allows lay users to build and use these models. For instance, leveraging the rich knowledge of LLMs will allow nonexpert developers to easily generate effective, context-relevant search spaces and associated hyperparameters for data augmentation.  In addition, intuitive user interfaces based on LLMs will enable lay users to simply specify natural language instructions to configure various settings of the resulting model. These frameworks, through the LMM interfaces, can also provide useful feedback and suggestions, as well as explanations about the internal mechanisms and operation of the model.

	\item The automation of data augmentation is an extremely computationally intensive process. Given that for a dataset several augmentations can be combined in different ways (e.g., varying numbers of transformation operations and different ordering of their application) and each transformation operation can be applied with infinitely wide range of intensities, the number of possible augmentations is effectively unlimited. In order to establish the effectiveness of any new augmentation using appropriate search strategies, it is usually necessary to conduct exhaustive training and subsequently validate the performance on a proxy task or the target task. This procedure is a combinatorial problem and, using current approaches, in some cases it is impractical to arrive at an optimal solution within a practical time frame. A potentially effective workaround is to leverage large language model prompts to guide the search process towards better solutions. Such a mechanism could also provide a means of high-level interaction that allows developers' priorities and constraints   (e.g., infrastructure limitations, economic constraints, time laxity, etc.) to be factored in the search process.  The performance evaluation step could also benefit from such new techniques by leveraging real-time feedback from LLMs.
		
	\item One of the major problems of machine learning tasks that require data augmentation is imbalanced data—a situation where some classes are underrepresented while others are overrepresented. This results in a problem where predictive performance on minority classes is severely compromised. Even though there are data augmentation-based workarounds specifically designed to overcome this problem (i.e., to balance imbalanced data), automated data augmentation methods have not yet been extended to this domain. Future research is expected to produce dedicated AutoML frameworks specifically aimed at balancing imbalance training data. 
	
	\item Currently, the automated data augmentation process is based on creating different variations of the input data and combing them in a random manner. However, it is known that the effectiveness of the data augmentation greatly depends on the order of augmentation. We expect future research to provide better theoretical grounding relating to the (general) ordering of augmentation operations for specific data types and tasks. Such an approach may rely on knowledge of the representative quality of the data with respect to the given task as various transformations are applied in sequence. Alternatively , principled formulations may be devised to quantify in advanced the effect of applying given transformations in specific sequences.
	
	\item In manual data augmentation processes, the human expert relies on intuition and domain expertise to determine the most suitable augmentations and their strengths. This greatly reduces the number of artificial samples that need to be created. It also increases robustness and reliability since the most important missing samples will logically be included in the augmented data. Automated augmentation methods, on the other hand, typically create a large number of redundant samples. In some situations, this can adversely harm performance and reliability. Overcoming this challenge requires the incorporation of context knowledge in automated data augmentation schemes. Context-aware automated data augmentation strategies would become vital in the near future as the requirements for machine learning systems are continuously been pushed to the limits. This objective may be achieved by introducing additional, domain-specific optimizations to force the generated data to align with real-world constraints . To accomplish this, logical rules or domain knowledge represented in any appropriate form may be utilized. 
	
	\item While conventional data augmentation strategies rely on constructing dataset- and task-specific solutions, automatically learned policies have been shown (e.g., \cite{cubuk2019autoaugment}) to be generally more transferable to new datasets. This property could potentially be exploited to provide further insights towards developing generic augmentations strategies for a wider range of deep learning tasks that are independent of datasets.  Some interesting results \cite{cao2023autotransfer} in this direction have already been reported. Future line of research will involve devising mechanisms to optimize models based on their transferability. This will enable the underlying models to encode more  transferable architectureual features and hyperparameter settings, thereby generating task- and dataset-agnostic AutoML frameworks. 
	
\end{itemize}

\section{Conclusion}
In this work we present a comprehensive survey of data augmentation methods based on AutoML techniques. We discuss different ways of realizing data augmentation using AutoML approaches. In particular, we cover data manipulation, data integration and data synthesis techniques. We also consider approaches for accomplishing all the subtasks of the data augmentation process: search space construction, model and hyperparameter optimization, and evaluation of intermediate solutions. Importantly, we provide a thorough discussion of the performance of automated data augmentation methods. In this regard, the performance of automated data augmentation methods are compared with results based on state-of-the-art classical approaches.  The results show that automated data augmentaation methods are currently superior to classical approaches in terms of predictive performance. Their main drawback, however, is their relatively high complexity of models and the enormous computaional requirements. Because of their enormous performance advantage, the automated methods are expected to improve further and assume an even more dominant role in large-scale data augmentation tasks.

\appendices



\ifCLASSOPTIONcaptionsoff
  \newpage
\fi

\end{document}